\documentclass[pdflatex,sn-nature]{sn-jnl}

\usepackage{graphicx}%
\usepackage{multirow}%
\usepackage{amsmath,amssymb,amsfonts}%
\usepackage{amsthm}%
\usepackage{mathrsfs}%
\usepackage[title]{appendix}%
\usepackage{xcolor}%
\usepackage{textcomp}%
\usepackage{manyfoot}%
\usepackage{booktabs}%
\usepackage{longtable}%
\usepackage{algorithm}%
\usepackage{algorithmicx}%
\usepackage{algpseudocode}%
\usepackage{listings}%

\usepackage{siunitx}
\usepackage[capitalize,nameinlink]{cleveref}
\usepackage{tabularx}
\usepackage{tikz}
\usepackage{pgfplots}
\pgfplotsset{compat=1.18}

\usepackage{algorithm}
\usepackage{algpseudocode}

\usepackage{newfloat}
\usepackage{caption}
\DeclareFloatingEnvironment[
  fileext   = lov,
  listname  = {List of Videos},
  name      = Vid.,        
  placement = tbhp,
]{video}
\def\fnum@video{{\bfseries\videoname\space\thevideo}}%
\def\VideoName{video}%
\makeatletter
\long\def\@makecaption#1#2{%
    \ifx\FigName\@captype
      \vskip\abovecaptionskip
        \@figurecaption{#1}{#2}
    \else\ifx\VideoName\@captype
      \vskip\abovecaptionskip
        \@figurecaption{#1}{#2}
    \else
        \@tablecaption{#1}{#2}
      \vskip\belowcaptionskip
    \fi\fi%
}
\makeatother
\crefname{video}{Vid.}{Vid.}
\Crefname{video}{Vid.}{Vid.}

\theoremstyle{thmstyleone}%
\theoremstyle{thmstyletwo}%

\theoremstyle{thmstylethree}%

\begin{document}

\title[PREVENT-JACK]{PREVENT-JACK: Context Steering for Swarms of Long Heavy Articulated Vehicles}

\author*[1,2]{\fnm{Adrian} \sur{Baruck}}\email{adrian.schoennagel@ovgu.de}

\author[1]{\fnm{Michael} \sur{Dub\'{e}}}\email{michael.dube@ovgu.de}
\equalcont{These authors contributed equally to this work.}

\author[2]{\fnm{Christoph} \sur{Steup}}\email{christoph.steup@ivi.fraunhofer.de}
\equalcont{These authors contributed equally to this work.}

\author[1,2]{\fnm{Sanaz} \sur{Mostaghim}}\email{sanaz.mostaghim@ovgu.de}

\affil*[1]{\orgdiv{Chair of Computational Intelligence}, \orgname{Otto-von-Guericke-University}, \orgaddress{\street{Universitätsplatz 2}, \city{Magdeburg}, \postcode{39106}, \country{Germany}}}

\affil[2]{\orgname{Fraunhofer Institute for Transportation and Infrastructure Systems IVI}, \orgaddress{\street{Zeunerstr. 38}, \city{Dresden}, \postcode{01069}, \country{Germany}}}


\abstract{%
In this paper, we aim to extend the traditional point-mass-like robot representation in swarm robotics and instead study a swarm of long Heavy Articulated Vehicles (HAVs). 
HAVs are kinematically constrained, elongated, and articulated, introducing unique challenges. Local, decentralized coordination of these vehicles is motivated by many real-world applications. 
Our approach, Prevent-Jack, introduces the sparsely covered context steering framework in robotics. It fuses six local behaviors, providing guarantees against jackknifing and collisions at the cost of potential dead- and livelocks, tested for vehicles with up to ten trailers.
We highlight the importance of the Evade Attraction behavior for deadlock prevention using a parameter study, and use 15,000 simulations to evaluate the swarm performance. Our extensive experiments and the results show that both the dead- and livelocks occur more frequently in larger swarms and denser scenarios, affecting a peak average of \SI{27}{\percent}/\SI{31}{\percent} of vehicles. We observe that larger swarms exhibit increased waiting, while smaller swarms show increased evasion.
}

\keywords{Context Steering, Swarm Robotics, Truck-Trailer, Heavy Articulated Vehicle}



\maketitle


\section{Introduction}

Heavy Articulated Vehicles (HAVs) are used in many applications such as logistics, agriculture, mining, and transportation. These domains particularly benefit from decentralized coordination, as it enables reactive, scalable, and human-like interactions between machines/robots and their environment. Relevant platforms include trucks with trailers in logistics centers, where yard-wide trailer maneuvering could be automated; remote mining and agricultural operations such as harvesting or simultaneous tillage and planting, where trailer-like tools and implements are moved across large areas, often with multiple robots sharing a workspace; and articulated buses, both in transit and in depots, where automation could help address labor shortages. Further examples include moving goods with Automated Ground Vehicles (AGVs) with trailers, such as airport baggage handling, trackless tourist trains, and manufacturing logistics. Despite this diversity of real-world use cases, decentralized, purely reactive approaches to coordinating a fleet of HAVs have, to the best of our knowledge, not yet been addressed in the literature.

In this paper, our main goal is to study a fleet of HAVs in the context of swarm robotics. 
Most existing models in swarm robotics represent robots as point masses or simple geometric bodies, such as discs or spheres, subject to basic kinematic constraints (e.g., single-integrator or differential-drive motion models). In contrast, HAVs are long vehicles towing one or more passive trailers. This feature leads to a fundamentally different shape and complex motion model, which cannot be captured by standard abstractions.

The complexity in such systems arises mainly from the hitch articulation and overall body length, which introduce two safety-critical challenges: (i) preventing \textit{jackknifing}, that is, infeasible hitch angles that cause self-collision, and (ii) developing space-efficient mutual collision-prevention strategies for swarms of elongated, non-spherical robots.

Our previous work, Avoid-Jack \cite{schonnagelAVOIDJACKAvoidanceJackknifing2025} introduced attraction-repulsion-based swarm behaviors for HAVs, providing an initial exploration of this domain. However, the complexity of collision avoidance substantially scales with the swarm size, demanding more sophisticated mechanisms for reconciling multiple competing objectives.
In this paper, we address this limitation using \textit{context steering}, a method well-suited for multi-behavior coordination. Context steering is originally studied in computer games \cite{frayContextSteeringBehaviorDriven2019}. We propose novel context steering behaviors tailored for HAV swarms that aim to provide provable guarantees against jackknifing and collisions. For this purpose, we propose \textit{Prevent-Jack} as a decentralized, purely reactive, consensus-free control approach for swarms of HAVs.

To validate our approach, we use a scalable kinematic simulation framework, whose realism we confirm with physics-based simulations in Gazebo. We fine-tune Prevent-Jack's parameters in a simulation study and compare its performance to Avoid-Jack. We then extensively analyze swarm performance in terms of deadlock and livelock rates, average vehicle speed, and path deviation. These metrics are assessed across various swarm sizes and collision densities in randomized scenarios, with a total of 15{,}000 simulations.

The contributions of this paper are as follows: 

\textbf{Preventing Jackknifing:} We introduce an original control algorithm for decentralized fleets of HAVs that 
provides guarantees against jackknifing and inter-vehicle collisions. These guarantees come at the cost of potential deadlocks and livelocks when large swarms operate in dense environments, which we analyze and discuss.

\textbf{Kinematic Constraints:} We extend the typical robot abstractions used in swarm robotics to encompass highly kinematically constrained vehicles, thus narrowing the gap between idealized swarm models and the requirements of real-world articulated platforms.

\textbf{Context Steering:} We transfer \emph{context steering}~\cite{frayContextSteeringBehaviorDriven2019}, a multi-behavior merging strategy used in computer games, to robotics and specifically to the control of HAV swarms. Unlike standard behavior fusion approaches that combine only the resulting motion decisions, context steering propagates the context in which each behavior would make its decision into the merging process. As a result, the decision layer merges decision contexts rather than bare velocity vectors. This enables human-understandable design of each goal objective and drastically reduces the need for fine-tuning behavior-merging parameters.


\section{Related Work}
\label{sec:sota}
Multi-robot coordination has been an active research area for several decades, with approaches broadly categorized into centralized and decentralized schemes. In centralized methods, the trajectories of all agents are planned and coordinated by a central entity. This has been extensively studied for robots with simple geometries and kinematics \cite{capPrioritizedPlanningAlgorithms2015a,sternMultiAgentPathfindingDefinitions2019} and partially extended to more complex vehicles \cite{kepplerPrioritizedMultiRobotVelocity2020}. Due to the high computational complexity, most approaches rely on path–velocity decomposition~\cite{kantEfficientTrajectoryPlanning1986}, thereby shifting the responsibility for avoiding jackknifing to the underlying path planner. In multi-robot multi-path planning (MRMP), joint trajectory planning methods such as K-CBS \cite{kottingerKCBS} employ centralized off-robot planning combined with problem decomposition. However, these methods assume complete a priori knowledge of the environment and agent set and compute full trajectories before execution, requiring recomputation upon unforeseen circumstances.

Decentralized approaches, in contrast, offer increased flexibility, adaptability to dynamic environments, and scalability by relying on local decision-making, where each robot plans based on partial information. This makes it substantially more challenging to enforce articulation and collision constraints.

\paragraph{Predictive, Decentralized Approaches}

Existing distributed model predictive control (MPC) formulations incorporate trailers only for specific maneuvers such as lane changes \cite{dmpc_truck,dmpc_truck_trailer_platooning}. Likewise, the literature on distributed control for multi-robot systems (MRS) typically targets specific scenarios, for example, leader–follower behaviors \cite{mrs_distributed_control}, rather than general multi-agent navigation with articulated vehicles.

A prominent class of predictive decentralized collision avoidance methods, including RVO~\cite{vandenBergRVO08} and ORCA~\cite{vandenBergORCA11}, operates via linear programming over velocity half-planes and is well established for holonomic agents. Nonholonomic extensions include NH-ORCA~\cite{alonso-moraOptimalReciprocalCollision2013} and curvature-constrained variants, e.g., \cite{rufliReciprocalCollisionAvoidance2013,bareissGeneralizedReciprocalCollision2015}. Articulated multi-trailer systems, however, introduce configuration-dependent, non-convex feasible velocity sets and jackknife constraints that are clashing with the ORCA half-plane formulation. To our knowledge, no direct extension of ORCA to car-trailer systems exists; the closest approach couples ORCA with a downstream MPC enforcing differential-drive kinematic limitations \cite{maoOrcaMPC20}, which might be extendable to HAVs. 

Other predictive approaches include Monte Carlo Tree Search (MCTS) variants, e.g., \cite{bone23MonteCarlo}, and rolling horizon evolutionary algorithms, cf. \cite{perez13RHEA,gaina22RHEA}. To our knowledge, these approaches are not extended to HAVs yet. 

\paragraph{Reactive, Decentralized Approaches}

At the extreme end of decentralization lies purely reactive planning. Within this paradigm, swarm robotics has predominantly focused on agents with simple kinematics \cite{hamannSwarmRoboticsFormal2018,diasSwarmRoboticsPerspective2021,engelbrechtFundamentalsComputationalSwarm2006}. Extensions to non-holonomic systems, such as cars \cite{kumarLyapunovBasedControlSwarm2015} and fixed-wing aircraft \cite{hauertReynoldsFlockingReality2011}, introduce basic motion constraints but typically rely on simplified collision-avoidance strategies and do not capture more complex kinematic behaviors such as articulation limits.

Multi-behavior decision-making frameworks, such as context steering \cite{frayContextSteeringBehaviorDriven2019}, offer promising advantages over single-policy approaches, that is, approaches where each behavior generates a singular preferred action. In context steering, each behavior (e.g., goal seeking, obstacle avoidance, cohesion) generates a context map that assigns measures of desirability or risk to candidate motion commands (typically headings or velocities). These maps are then aggregated, and at each control step the action with maximal resulting desirability under acceptable risk is selected, enabling fine-grained arbitration between competing behaviors. To the best of our knowledge, context steering has not yet been applied to robotics in peer-reviewed publications. The only reported robotics implementations are two M.Sc. theses on quadrotor swarms \cite{stein_multikriteriell_2018,thoms_anwendung_2021}. In both theses, the quadrotors are modeled as point-like agents in 2D or 3D space rather than highly constrained platforms such as HAVs.

\paragraph{Contribution}

Our focus is on purely reactive, decentralized planning for robots with complex kinematics. To the best of our knowledge, only our previous work, Avoid-Jack~\cite{schonnagelAVOIDJACKAvoidanceJackknifing2025}, explicitly addresses this problem, using attraction–repulsion-based behaviors. However, Avoid-Jack is limited to two agents, provides only restricted handling of multiple competing constraints, and does not guarantee the absence of jackknifing or collisions. In this paper, we therefore propose a context steering approach tailored to articulated multi-robot multi-trailer systems that is designed to overcome these limitations. This approach will be referred to as Prevent-Jack.

\section{Problem Description \& Vehicle Model}
\label{sec:problem}

We originally introduced the problem studied in this paper in \cite{schonnagelAVOIDJACKAvoidanceJackknifing2025}. In summary, the swarm consists of $N_H$ HAVs, where each HAV $i\in\{0,\dots,N_H-1\}$ is an Ackermann-steered truck towing $N_i\in\mathbb{Z}^+$ passive trailers. In a swarm of HAVs, two new major challenges arise. First, vehicles may enter invalid state-space configurations through over-articulation (that is, by exceeding inter-segment joint limits), a phenomenon commonly referred to as \textit{jackknifing} or \textit{self-collision}. Second, conventional circular footprint approximations, while geometrically safe, become increasingly conservative as vehicle articulation grows, resulting in a substantial overestimation of the space occupied by each robot.

In the present work, we address the jackknifing problem as a necessary step towards enabling reliable swarm operation of HAVs. Circular footprint approximations are retained as a safe, if conservative, baseline for collision avoidance. The remainder of this section is organized as follows: we first describe the vehicle morphology and its kinematic model, then formally define the jackknifing and collision constraints, and finally introduce the scenario under consideration. We summarize all introduced symbols at the end of this paper.  

\begin{video}[hbt]
    \centering
    \vspace{1.3cm} 
    \includegraphics{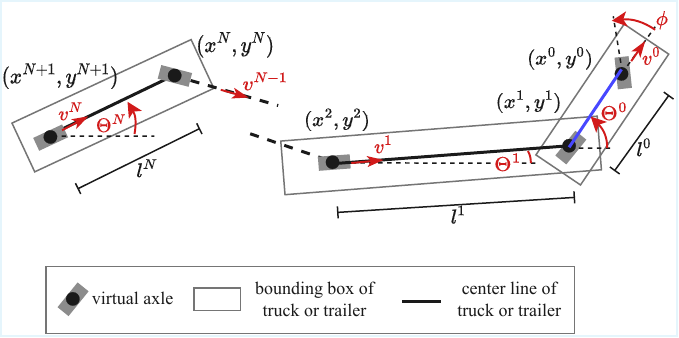} 
    \vspace{1.3cm} 
    \caption{Ackermann truck–trailer model for HAV $i$. Truck (blue) and first trailer (black) on the right; final trailer $N_i$ on the left. Intermediate trailers omitted (thick dashed). Subscript $i$ is omitted from variables for clarity. In the PDF version of this article, please refer to \textbf{vid1\_kinematics.mp4} in the accompanying archive to view the video.}
    \label{fig:kinematics}
\end{video}

\subsection{Kinematic Model}
\label{sec:problem:kinematics}

Each HAV $i$ is represented as a truck-trailer system comprising $N_i$ trailers; cf. \cref{fig:kinematics}. To simplify the model, we assume negligible vehicle width, on-axle hitching, and wheelbase equal to segment length. A \textit{virtual axle} is defined as the centroid of all corresponding physical axles. The truck's front and rear virtual axles are assigned indices $k=0$ and $k=1$, respectively, while each trailer 
$j\in[1,N_i]\subset\mathbb{N}$ carries a single rear virtual axle at index $k=j+1$. The spatial position of the $k^{\text{th}}$ virtual axle is denoted $(x_i^k, y_i^k)$, and the headings of the truck and trailer $j$ are written as $\Theta_i^0$ and $\Theta_i^j$, respectively. The wheelbases of the truck and trailer $j$ are $l_i^0$ and $l_i^j$. A schematic illustration of this configuration is provided in \cref{fig:kinematics}.

The motion of each HAV is governed by two control inputs applied to the truck: its longitudinal speed $v_i^0 \geq 0$ and steering angle $\phi_i \in [-\phi_i^{\text{max}}, \phi_i^{\text{max}}]$. The truck's heading $\Theta_i^0$ evolves with speed $v_i^0$, and each trailer $j$ inherits a speed $v_i^j$ propagated via the tugging forces from its predecessor. Note that vehicles are restricted to forward driving at the current stage due to HAVs unstable internal dynamics and coupled nonlinear terms present when reversing \cite{Salamah23TruckReversing}.   
Building on the on-axle hitching assumption, we adopt the kinematic differential equations from \cite{sordalenConversionKinematicsCar1993}, expressed as:
\begin{align}
    \begin{pmatrix}
        {\dot{x}_i^1}\\
        {\dot{y}_i^1}\\
        {\dot{\Theta}_i^0}
    \end{pmatrix} = \begin{pmatrix}
        {v_i^0 \cdot \cos(\Theta_i^0)}\\
        {v_i^0 \cdot \sin(\Theta_i^0)}\\
        {v_i^0/\mathrm{l_i^0} \cdot \tan(\phi_i)}
    \end{pmatrix}
    \label{eq:kinematics}
\end{align}
\begin{align}
    \forall j\in[1,N_i]:
    \begin{pmatrix}
        {\dot{\Theta}_i^j}\\
        {v_i^j}
    \end{pmatrix} = \begin{pmatrix}
        {-\frac{v_i^{j-1}}{\mathrm{l_i^j}} \cdot \sin(\Theta_i^{j}-\Theta_i^{j-1})}\\
        {v_i^{j-1} \cdot \cos(\Theta_i^{j}-\Theta_i^{j-1})}
    \end{pmatrix} 
    \label{eq:kinematics-trailers}
\end{align}

\subsection{Jackknifing}
\label{sec:problem:jackknifing}

A critical failure mode in articulated vehicles is jackknifing, which arises when the relative angle between two consecutive segments exceeds a permissible threshold, potentially causing irreversible structural damage. We define the articulation angle between trailer $j$ and its preceding segment as $\delta_i^j = \Theta_i^j - \Theta_i^{j-1}$, for $j = 1, \dots, N_i$. Applying standard trigonometric identities, the angular inequality constraint \eqref{eq:delta_lim} can be equivalently expressed as \eqref{eq:cos_delta_lim}. While we adopt a limit of \SI{90}{\degree} ($\pi/2$) throughout this work, the formulation generalises to any angle within $(0,\,\pi]$:
\begin{align}
    \forall& j\in[1,N_i]:&  \pi/2 &\geq \Big|\delta_i^j\Big| \label{eq:delta_lim}\\
    \forall& j\in[1,N_i]:&  0 &\leq \cos{\bigl(\delta_i^j\bigr)} \label{eq:cos_delta_lim}
\end{align}

\subsection{Mutual Collision Avoidance}
\label{sec:problem:collision}

Each HAV $i$ is enclosed by a bounding circle centred at its truck's rear axle $(x_i^1,\,y_i^1)$ with a \textit{collision radius} $d_i = \max\bigl\{l_i^0, \sum_{j=1}^{N_i} l_i^j\bigr\}$, referred to as the vehicle's \textit{footprint}. The combined \textit{collision distance} between HAVs $i$ and $h$ is then $d_{i,h} = d_i + d_h$. A \textit{potential collision} is flagged whenever these bounding circles overlap, as expressed in \eqref{eq:potential-collision}. For higher geometric fidelity, each HAV $i$ may alternatively be represented as a polygonal chain 
$P_i$ connecting its successive axles; an \textit{actual collision} between HAVs $i$ and $h$ is then declared when these chains intersect, as stated in \eqref{eq:collision}. In this work, we will solely utilize the circular approximations of the potential collision. 
\begin{align}
    \bigl\lVert 
        (x_i^1,\,y_i^1)^{\top} - (x_h^1,\,y_h^1)^{\top} 
    \bigr\rVert_2 
    &\leq d_{i,h}
    \label{eq:potential-collision} \\
    P_i \cap P_h &\neq \emptyset
    \label{eq:collision}
\end{align}

\subsection{Scenario}
\label{sec:problem:scenario}

We consider a swarm of $N_H$ HAVs navigating a two-dimensional toroidal environment with no known obstacles, such as to limit available space while focusing on intervehicle interaction without influence from static obstacles. Each vehicle $i$ is initialized at a \textit{start pose} $\mathcal{P}_i^S = \{p_{x,i}^S,\, p_{y,i}^S,\, p_{\Theta,i}^S\}$, which specifies the position $(x_i^1,\,y_i^1)$ and heading $\Theta_i^0$ of the truck's rear axle, with all trailers assumed to be fully aligned (zero articulation). Each vehicle is subsequently assigned a sequence of \textit{goal poses} $\mathcal{P}_i^G = \{p_{x,i}^G,\, p_{y,i}^G,\, p_{\Theta,i}^G\}$ for the same axle, wherein trailer articulation is permitted subject only to the jackknifing constraint \eqref{eq:cos_delta_lim}.

All start and goal poses are selected such that no pair of vehicles $(i,\,h) \in \{1,\dots,N_H\}^2$, $i \neq h$, satisfies the collision condition \eqref{eq:potential-collision} at initialization or upon reaching their respective goals.


\section{Contribution: PREVENT-JACK}
\label{sec:prevent-jack}

In this paper, we utilize context steering \cite{frayContextSteeringBehaviorDriven2019} by developing a total of six behaviors. Each behavior generates a \textit{context map} in each time step, which is either of type \textit{danger} or \textit{interest}.
In this work, context maps are two-dimensional structures as depicted in \cref{fig:context-steering}. They comprise discrete \textit{actions}, each representing a velocity-steering pair $(v_i^0,\phi_i)$ for HAV $i$ to be executed by itself over the next time step $\Delta_t$. Depending on the behavior's type, it assigns either a danger or interest value $\in[0,1]$ to every action, quantifying its risk or desirability, respectively.

\begin{video}[hbt]
    \centering
    \includegraphics[width=1\linewidth]{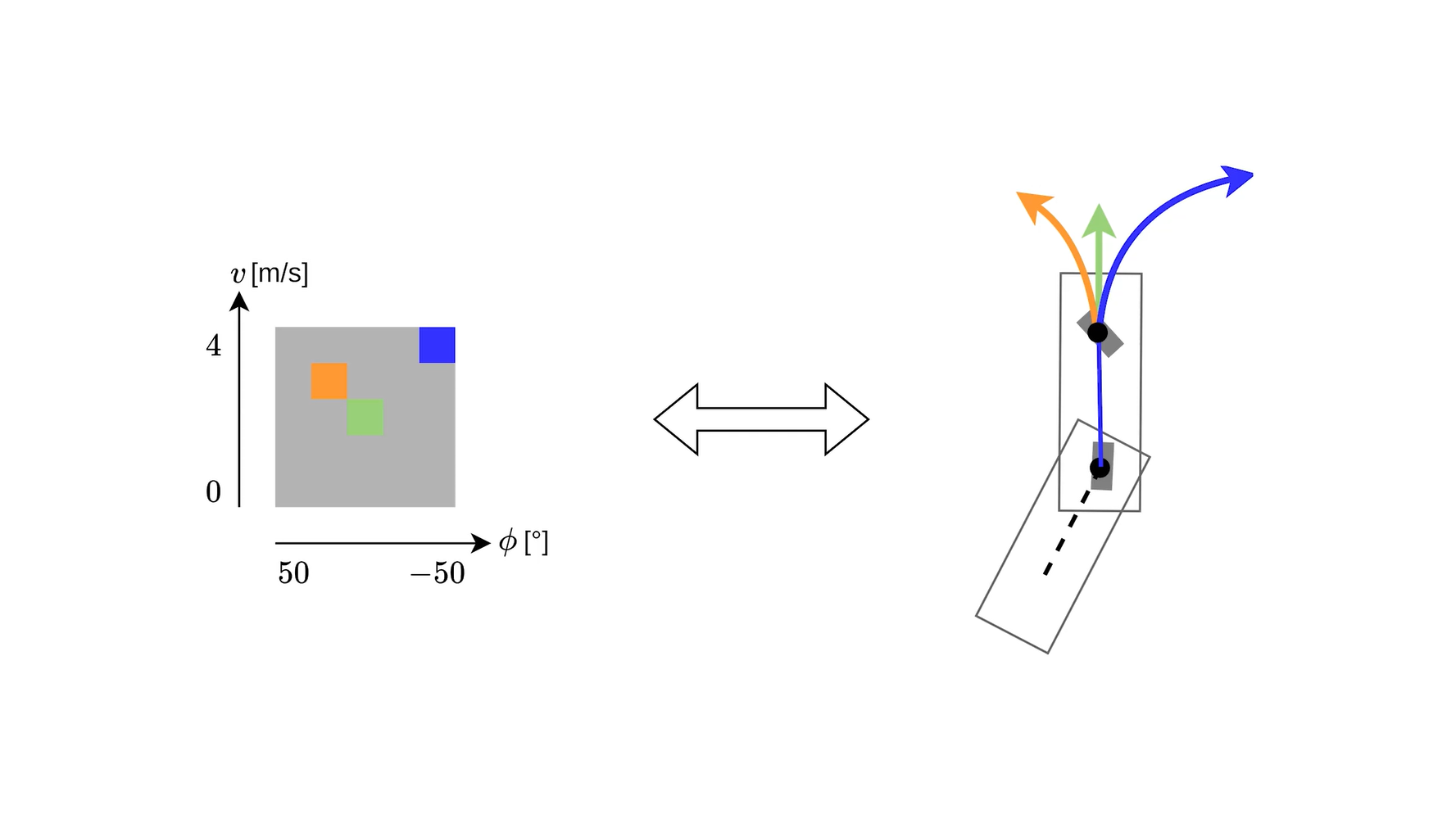}
    \caption{Context steering explained. Context maps are 2D structures of discrete actions the robot can take. After merging the context maps, the action with the highest desirability under acceptable risk is executed. In the PDF version of this article, please refer to \textbf{vid2\_context\_steering.mp4} in the accompanying archive to view the video.}
    \label{fig:context-steering}
\end{video}

The action to be executed is determined using \cref{alg:context-map-merging}. We first merge all danger maps via element-wise maximum and apply an epsilon constraint threshold, producing a binary \textit{block mask} identifying prohibited actions. The interest maps are then combined through a weighted sum and filtered by the block mask, setting blocked actions to zero interest. Thereby, danger behaviors enforce constraints to be held.

\begin{algorithm}
    \caption{Context map merging and action selection. Subscript $i$ omitted from variables for clarity. 
    The method \textproc{Upsample(values, $s_1, s_2$)} creates a 2D interpolation of values using the scale factors $s_1$ for the first axis and $s_2$ for the second axis.}
    \label{alg:context-map-merging}
    \begin{algorithmic}[hbt]
        \Require $\mathcal{C}_I\gets\{C_I^1,C_I^2,\dots\}$ \Comment{Set of all interest context maps}
        \Require $\{w_1, w_2, \dots\}$ \Comment{Set of weights for each interest context map}
        \Require $\mathcal{C}_D\gets\{C_D^1, C_D^2,\dots\}$ \Comment{Set of all danger context maps}
        \Require $\epsilon_D=0.1$ \Comment{Danger Threshold}
        \Require $V_{n_v}\gets\{v_m\}_{m=1}^{n_v}$, $\Phi_{n_\phi}\gets\{\phi_n\}_{n=1}^{n_\phi}$ \Comment{Discrete context-map grid in $v$ and $\phi$}
        \Require $v^{\min}, v^{\max}, \phi^{\min}, \phi^{\max}$ \Comment{Dynamic limits}
        \Require $n_v^\text{int}\gets 20, n_\phi^\text{int}\gets 40$ \Comment{Interpolation resolution in $v$ and $\phi$}
        \Ensure $(v^0,\phi)$ \Comment{Executable Action}
        \State \textbf{// Danger aggregation and masking}
        \State Block Mask $B \gets \bigvee_k \bigl(C_D^k > \epsilon_D\bigr)$
        \State Merged Interest $C_I^S \gets \sum_k w_k C_I^k$
        \State Filtered Interest $C_I^F \gets (1-B) \odot C_I^S$ \Comment{Element-wise product, i.e., 0 where $B=1$}
        \If{$B = \mathbf{1}$}
            \State $v^0 \gets 0$; \hspace{0.5em} $\phi \gets 0$ \Comment{All options dangerous, stand still}
            \State \Return $v^0,\phi$
        \EndIf
        \State \textbf{// Action selection by upsampling}
        \State $\hat{C}_I \gets$ \Call{Upsample}{$C_I^F,\; s_v,\, s_\phi$}
            \Comment{$s_v = \frac{n_v^\text{int}-1}{n_v-1}$,\; $s_\phi = \frac{n_\phi^\text{int}-1}{n_\phi-1}$}
        \State $p^*,q^* \gets \arg\max_{p,q}\; \hat{C}_{I,p,q}$
        \State $v^0 \gets v_{\min} + \dfrac{p^*-1}{n_v^\text{int}-1}(v_{\max}-v_{\min})$
        \State $\phi  \gets \phi_{\min} + \dfrac{q^*-1}{n_\phi^\text{int}-1}(\phi_{\max}-\phi_{\min})$
        \State \Return $v^0,\phi$
    \end{algorithmic}
\end{algorithm}

Following \cite{dockhornMultiObjectiveOptimizationDecisionMaking2021}, we interpolate the resulting context map to mitigate discretization effects. The discrete context map is upsampled to a higher resolution, from which the action corresponding to the highest interpolated interest value is selected for execution. Note that the interpolation naturally assigns lower interest values to the discretization neighborhood of blocked actions, preventing their selection. In the case where all actions are blocked, the HAV defaults to remaining stationary. Note that, while this fallback provides safety, it can itself cause deadlock cascades. An overview of the employed behaviors is presented in \cref{tab:behavior-overview}.
\begin{table}[ht]
    \centering
    \caption{Overview of behaviors.}
    \label{tab:behavior-overview}
    \begin{tabularx}{\linewidth}{>{\raggedright}p{.25\textwidth} X}
        \toprule
        Behavior & Description \\
        \midrule
        Dubins Goal Attraction & Attraction towards the goal based on Dubins path-planning and path-following controller.\\
        Jackknife: Prevention & Prevent HAV from entering jackknifed states.\\
        Jackknife: Straightening Attraction & Interest for HAV to keep straight, thereby making curling up unattractive.\\
        Collision: Prevention & Prevent HAV from colliding with others.\\
        Collision: Evade Attraction & Interest for HAV to keep a distance from others.\\
        Progress Attraction & Push HAV to keep moving if otherwise stalled.\\
        \bottomrule
    \end{tabularx}
\end{table}

Each HAV has access to its articulation state, the relative position and heading of its assigned goal, and the relative position and approximate collision radius of HAVs within its communication radius. In this work, we assume the communication radius to be $\geq2\cdot\max_h(d_h)+d_\text{evade}^B$, i.e., each HAV can at least obtain information from all HAVs with less separation gap than the upper bound of the evade attraction behavior $d_\text{evade}^B$; cf. \cref{sec:behavior:collisions}. In practical deployments, this information could be obtained via GNSS (Global Navigation Satellite System) combined with neighbor-to-neighbor communication or via decentralized localization and tracking methods such as SLAM~\cite{birdDvmSlam25,lajoieSwarmSlam24} or UVDAR~\cite{hornyaUvdar22,walterUvdar19}. The following sections detail how each behavior constructs its context map. Some behavior parameters and combination weights are chosen empirically, while others are tuned through the parameter study described in \cref{sec:eval:params}.

\subsection{Dubins Goal Attraction}
\label{sec:behavior:dubins}
%
Our previous work \cite{schonnagelAVOIDJACKAvoidanceJackknifing2025} introduced a Dubins-based path-planning approach for HAVs. Given a current pose $\{x_i^1,y_i^1,\Theta_i^0\}$ and goal pose $\{x_G,y_G,\Theta_G\}$, a connecting Dubins path \cite{dubinsCurvesMinimalLength1957} is computed, comprising circular arcs and at most one straight segment. The circular arc radius corresponds to the minimum stable turning radius $R_i^\text{min}$ of HAV $i$, defined by ${R_i^\text{min}}^2 = \sum_{j=0}^{N_i}(l_i^j)^2$~\cite{schonnagelAVOIDJACKAvoidanceJackknifing2025}. Note that this planner is purely geometric and does not account for obstacles.

In \cite{schonnagelAVOIDJACKAvoidanceJackknifing2025} we observed frequent goal misses, motivating the enhanced method presented here. Previously, a new Dubins path was planned at every time step and the executable steering command was chosen as either full left or full right, depending on the direction of the first path segment. Continuous replanning alone, however, led to suboptimal path-tracking performance. In the present work, we instead leverage path-following control: the Dubins path is recomputed only when the HAV exceeds a maximum \textit{path deviation} threshold $e_P^\text{max}=\SI{0.8}{\meter}$. Otherwise, we apply the combined steering control law in \labelcref{eq:controller-total}, which integrates a pure-pursuit Ackermann controller $\phi_i^P$ (cf. \cite{sniderAutomaticSteeringMethods2009}) as feedforward control with a Stanley cross-track error controller $\phi_i^S$ \cite{thrunStanleyRobotThat2006} for error correction.

\begin{figure}[hbt]
    \centering
    {\includegraphics[]{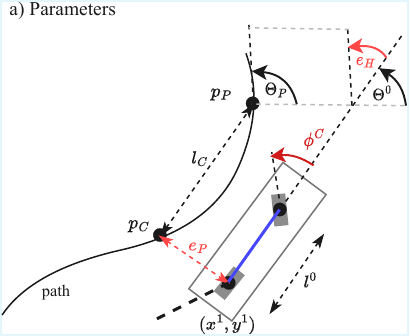}}
    \hfill
    \includegraphics[]{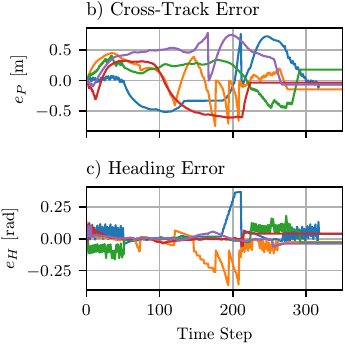}
    \caption{Parameters of the path-following controller. \textbf{a)} Visualization of parameter values in coordinate space. Subscript $i$ and trailers omitted for clarity; \textbf{b,c)} cross-track error $e_P$ and heading error $e_H$ over time for one random simulation with five HAVs (one per color). Jumps in errors indicate path replanning.}
    \label{fig:path-controller}
\end{figure}

The control computation proceeds as follows; see \cref{fig:path-controller}. We discretize the path into sample points and identify the point $p_C=(x^C,y^C)^T$ nearest to the truck's rear virtual axle. The \textit{lookahead point} $p_P=(x^P,y^P)^T$ with heading $\Theta^P$ is then determined by advancing a \textit{lookahead distance} $l_C = f_C \cdot l_i^0, f_C\in\mathbb{R}^{+}$ along the path. This yields the \textit{heading error} $e_H$ via \labelcref{eq:controller-heading-error} and the \textit{cross-track error} $e_P$ via \labelcref{eq:contoller-corss-track-error}, i.e., the angular deviation and Euclidean distance between the HAV and the closest and lookahead point, respectively. A visualization of the errors over time is depicted in \cref{fig:path-controller} for a random simulation with five HAVs. Jumps in the errors occur when paths are replanned due to evasive maneuvers; see \cref{sec:behavior:collisions}.  
\begin{align}
    e_H &= \Theta^P - \Theta_i^0 \label{eq:controller-heading-error} \\
    e_P &= \| (x_i^1,y_i^1)^T - (x^C,y^C)^T  \|_2 \label{eq:contoller-corss-track-error} 
\end{align}

The pure-pursuit component $\phi_i^P$ follows from \labelcref{eq:controller-pursuit}, while the Stanley component $\phi_i^S$ is computed according to \labelcref{eq:controller-stanley}, where $k_e \in\mathbb{R}$ represents the \textit{cross-track error gain}. Following the tuning guidance in \cite{sniderAutomaticSteeringMethods2009}, we determined the parameter values $f_C=0.2$ and $k_e = 2$ through empirical simulation studies.
\begin{align}
    \phi_i^P &= \arctan\bigl(2 l_i^0 \cdot e_H \big/ l_C\bigr) \label{eq:controller-pursuit} \\
    \phi_i^S &= \arctan\bigl(k_e \cdot e_P \big/ v_i^\text{max}\bigr) \label{eq:controller-stanley}\\
    \phi_i^C &= \phi_i^P + \phi_i^S \label{eq:controller-total}
\end{align}

This yields the optimal action $(\phi_i^C, v_i^\text{max})$ for the goal attraction behavior. To generate the corresponding context map, we define a two-dimensional Gaussian in the action space, centered at $(\phi_i^C, v_i^\text{max})$ with standard deviations $\sigma_\phi=\SI{1}{\radian}$ and $\sigma_v=\SI{2}{\meter\per\second}$, normalized to unit peak height. 
This Gaussian shaping of the context map yields a clear preference for the optimal action while still assigning moderate interest to actions with the same steering angle at lower velocities and to nearby steering angles. Consequently, if all actions with the exact optimal steering angle were blocked by safety constraints, actions with similar steering angles still provide meaningful progress toward the goal. The introduction of this path-following controller substantially improves path tracking and goal achievement in the absence of disturbances, as demonstrated in \cref{sec:eval:compare}.

\subsection{Jackknifing}

Jackknifing, i.e., too high trailer articulation, is a major safety concern for HAVs. Therefore, we introduce a danger behavior to prevent imminent constraint violation and pair it with an interest behavior that guides the vehicle towards reduced articulation. 

\subsubsection{Danger: Jackknife Prevention}
\label{sec:behavior:jack-prevent}

This behavior evaluates the kinematic feasibility of candidate actions. For each discrete action in the context map, we perform a forward simulation over a single time step using the HAV kinematic model described in \cite{schonnagelAVOIDJACKAvoidanceJackknifing2025}. Actions that induce a jackknifed configuration are assigned a danger value of 1; all others receive a danger value of 0.

While this mechanism successfully blocks immediate jackknifing maneuvers, it does not provide repulsion from states that lead to kinematic entrapment—configurations where all available, non-zero-velocity actions would result in jackknifing. To address this limitation, we introduce a complementary interest-based behavior that actively guides the vehicle away from highly articulated configurations.

\subsubsection{Interest: Straightening Attraction}
\label{sec:behavior:straightening}

\begin{figure}[hbt]
    \centering
    \begin{tikzpicture}
        \begin{axis}[
            axis lines = left,
            xlabel={Articulation Angle, $\delta^j$ [$\pi$]},
            ylabel={$f_{I_J}$ [1]},
            xmin=-1, xmax=1,
            ymin=0, ymax=2,
            clip=false,
            height=3.5cm,
            width=.8\linewidth
        ]
        \addplot [
            domain=-1:1, 
            samples=100, 
            color=teal,
        ]
        {1 + tanh(1 * (0.5 - 2 * cos(deg(pi*\x))))};
        
        \def\Xtwo{0.5}

        \addplot[gray] coordinates {(\Xtwo,0) (\Xtwo,1.46)};
        \node[above, darkgray] at (axis cs:\Xtwo+0.1,0) {$\delta_i^\text{max}$};

        \addplot[lightgray] coordinates {(-1,1) (1,1)};
        
        \end{axis}
    \end{tikzpicture}
    \caption{Function for straightening attraction per hitch-joint over articulation angle, see \labelcref{eq:straight_attract:sub}.}
    \label{fig:jackknife-weight}
\end{figure}
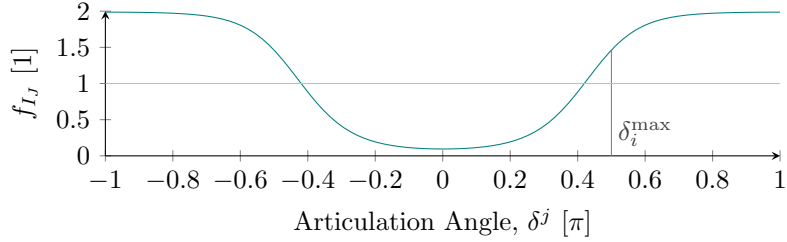

The Straightening Attraction behavior assigns interest to actions that reduce the HAV's articulation, that is, to driving straight. The interest values, and therefore the desire to pull straight, increase monotonically with the degree of articulation; see \labelcref{eq:straight_attract:sub} and \cref{fig:jackknife-weight}. This creates a gradient that guides the vehicle toward straighter configurations, enabling recovery from near-jackknife states and avoiding convergence toward kinematic entrapment. The function parameters were determined through empirical simulation studies. It is crucial, however, to decouple this interest from the hard danger constraint, as driving straight might not be feasible due to other constraints such as collision avoidance.  

The final, articulation-dependent interest function \labelcref{eq:straight_attract:main}, originally presented in \cite{schonnagelAVOIDJACKAvoidanceJackknifing2025} under “Jackknife Avoidance,” is applied exclusively to actions with zero steering angle ($\phi_i=0$), while others receive an interest of 0. Consequently, all context maps must include a discrete row corresponding to $\phi_i=0$ to enable this straightening mechanism. 
\begin{align}
    \label{eq:straight_attract:main}
    I_J &= \sum_{j=1}^{N_i} \Big[ \frac{1}{j^{0.2}} \cdot f_{I_J}(\delta_i^j) \Big] \\
    f_{I_J}(\delta_i^j) &= 1 + \tanh\bigl( 0.5 - 2 \cos \delta_i^j \bigr) \label{eq:straight_attract:sub}
\end{align}

\subsection{Mutual Collisions}
\label{sec:behavior:collisions}

Similar to jackknifing, preventing (mutual) collisions is an important safety constraint, needing to be enforced by a danger behavior. Similarly, a guiding interest to evade from others, while not enforcing immediate evasion, is needed as well.  

\begin{figure}[hbt]
    \centering
    \includegraphics[]{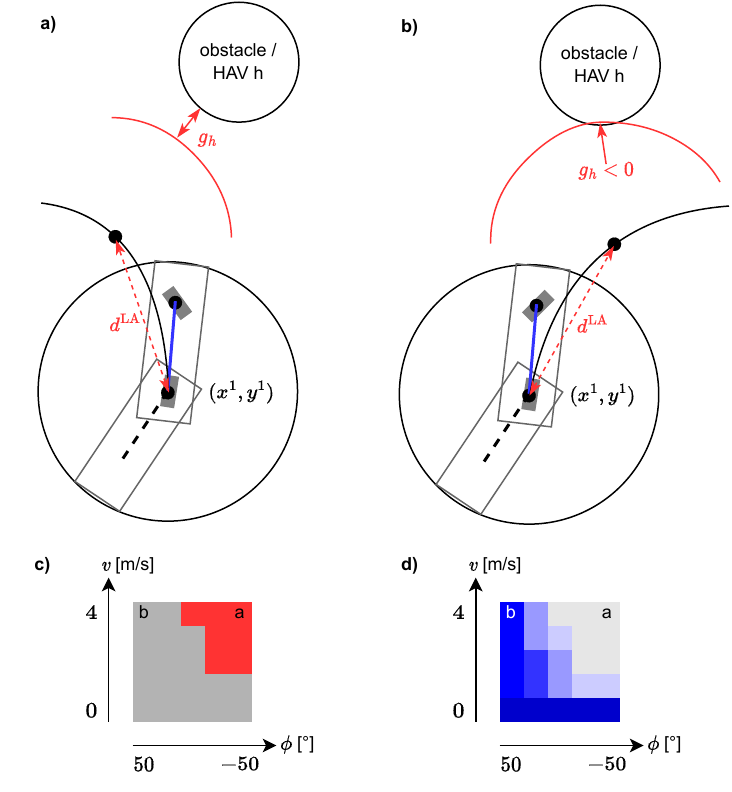}
    \caption{Mutual collision detection and parameters at an example obstacle. \textbf{a)} Evaluation of driving left; \textbf{b)} Evaluation of driving right in the same situation; \textbf{c)} Resulting danger context map in current step (red is danger); \textbf{d)} Resulting interest context map in current step (darker blue = higher interest).}
    \label{fig:mutual_collision_params}
\end{figure}

\Cref{fig:mutual_collision_params} shows an exemplary HAV $i$ approaching an obstacle HAV $h$. Both danger and interest behavior are based on trajectory simulation and geometric overlap detection. For each candidate action, we simulate HAV $i$'s trajectory using the kinematic model from \cite{schonnagelAVOIDJACKAvoidanceJackknifing2025} over a lookahead distance $d^\text{LA}$ and compute the remaining separation $g_h$ to HAV $h$. Subfigures a) and b) visualize the difference between a left and a right turn, while subfigures c) and d) depict the resulting context maps, respectively. 

\subsubsection{Danger: Collision Prevention}
\label{sec:behavior:col-rep}

For the danger behavior, we choose a lookahead distance $d_\text{coll}^\text{LA}=\SI{2}{\meter}$ and set the danger value to the sum of intersections (i.e., $g_h<0$) between HAV $i$ and all other HAVs $h$.

The \SI{2}{\meter} horizon is a key design choice. With articulated kinematics, steering inputs require distance before inducing meaningful heading change; while a single-step lookahead blocks collisions, it allows the HAV to approach obstacles too closely, leaving no room for evasive actions (see next subsection).

It is important to note that this behavior solely blocks collision-inducing actions but does not generate repulsive gradients. Consequently, a complementary attraction-based behavior is required to maintain separation distances from neighboring HAVs and therefore reduce the risk of deadlocks, as detailed in the following subsection.

\subsubsection{Interest: Evade Attraction}
\label{sec:behavior:evade}

The Evade Attraction behavior assigns higher interest values to actions that preserve greater distances from other vehicles.

For each candidate action, we simulate HAV $i$'s trajectory over a lookahead distance $d_\text{evade}^\text{LA} = \SI{8}{\meter}$ and evaluate the endpoint configuration. We compute the separation $g_h$ between HAV $i$'s endpoint and each neighboring HAV $h$ and apply a distance-dependent interest penalty. For a neighboring HAV with separation $g_h$, the penalty is computed as
\begin{align}
    p(g_h) = \begin{cases}
        1 & \text{, if } g_h<0\\
        \left(1 - \frac{g_h}{d_\text{evade}^B}\right)^{E_c} & \text{, elif } g_h < d_\text{evade}^B \\
        0 & \text{, otherwise}
    \end{cases}
    \label{eq:separation-penalty}
\end{align}
where $d_\text{evade}^B = \SI{10}{\meter}$ defines the distance threshold beyond which no penalty applies and $E_c = 4$ controls the steepness of the interest reduction. The final interest value for an action is then
\begin{align}
    I_\text{evade} = \max\left\{0, \Big(1 - \sum_{h} p(g_h)\Big)\right\}
    \label{eq:evade-interest}
\end{align}
where the sum is taken over all neighboring HAVs and the result is clipped to ensure non-negative interest values. 

The extended lookahead distance $d_\text{evade}^\text{LA} = \SI{8}{\meter}$ creates substantial spatial differentiation between action outcomes, enabling effective separation guidance. In merging with the other behaviors, the weight $w_\text{evade} = 2$ amplifies the behaviors influence in the final action selection. All parameters and their performance effects are systematically analyzed in the parameter study (see \cref{sec:eval:params}), with baseline values determined through preliminary simulation experiments.

\subsection{Progress Attraction}
\label{sec:behavior:progress}

Preliminary experiments revealed systematic deadlocks where opposing HAVs halt indefinitely, each waiting for the other to yield. To resolve such standoffs, we introduce a Progress Attraction behavior that incentivizes movement.

This behavior monitors HAV $i$'s velocity and increments a counter $n_\text{standstill}$ during standstill ($v_i^0 = 0$). The interest value increases periodically according to
\begin{align}
    I_\text{prog} = \left\lfloor \frac{n_\text{standstill}}{N_\text{prog}} \right\rfloor \cdot \Delta_{I_\text{prog}}
    \label{eq:progress-interest}
\end{align}
where $N_\text{prog} = 15$ defines the period between interest increments and $\Delta_{I_\text{prog}} = 0.15$ specifies the increment magnitude. This interest applies to all actions with $v>0$, creating increasing pressure to resume motion. Both $I_\text{prog}$ and $n_\text{standstill}$ reset to zero when movement occurs.

This behavior models driver impatience, with parameters influencing the trade-off between cautious waiting and assertive progress. Parameter values and their performance impact are evaluated in the parameter study (see \cref{sec:eval:params}), with baselines determined through preliminary experiments.

\subsection{Properties}
\label{sec:properties}

We summarize the core properties of the proposed Prevent-Jack algorithm below.

\subsubsection{Guarantees}

In the absence of inter-robot collision risk (for example, in single-robot deployments or in swarms with non-intersecting trajectories), jackknife prevention is guaranteed. This guarantee follows directly from the fact that Dubins Path Attraction exclusively generates reference paths with curvatures at or above the minimum stable turning radius, which are inherently jackknife-free~\cite{schonnagelAVOIDJACKAvoidanceJackknifing2025}. Residual deviations from the reference path, arising from tracking error, remain minor, and any vehicle state approaching a dangerous configuration is precluded by the Jackknife Prevention Behavior. Kinematic entrapment does not arise when the vehicle adheres to the Dubins path.

When robot trajectories intersect and evasive maneuvers become necessary, the absence of jackknifing and collisions is guaranteed, resulting in no damage to the vehicle. Although such maneuvers produce departures from the Dubins path, the Jackknife Prevention and Collision Prevention components jointly prohibit the execution of any unsafe actions. However, the algorithm does not guarantee solution completeness: deadlocks may arise when multiple robots meet simultaneously. While Evade Attraction and Straightening Attraction are designed to guide vehicles through such configurations, they operate without predictive modelling and therefore cannot ensure completeness. This constitutes an inherent limitation of the reactive paradigm underlying the presented approach.

\subsubsection{Communication Complexity}

Each robot emits a single, regular broadcast, yielding a total communication effort of $\mathcal{O}(N_H)$.

\subsubsection{Memory Complexity}

Each robot $i$ maintains the relative pose and size of all $N_H-1$ neighboring robots within its communication range and its own $N_i$ trailer angles, resulting in a worst-case memory complexity of $\mathcal{O}(N_H + N_i)$.

\subsubsection{Computational Complexity}

On a per-robot basis, Dubins Path Attraction and Progress Attraction Behaviors each operate in $\mathcal{O}(1)$; jackknife-related behaviors scale as $\mathcal{O}(N_i)$; collision-related behaviors scale as $\mathcal{O}(N_H)$; and behavior merging operates in $\mathcal{O}(1)$. The overall per-robot computational complexity is therefore $\mathcal{O}(N_H + N_i)$.

\section{Evaluation}
\label{sec:eval}

We evaluate Prevent-Jack's performance in kinematic and physics-based simulation environments. We first introduce the simulation frameworks, the scenario generation, and the used metrics in \cref{sec:eval:methodology}.
Then, the remainder of the evaluation is structured as follows: first, we tune the algorithm in a parameter study, described in \cref{sec:eval:params}. Then, \cref{sec:eval:gazebo} compares kinematic and physics-based simulations and verifies that the kinematic approximations are valid. Then, we compare the proposed Prevent-Jack to literature baselines, namely our Avoid-Jack, in \cref{sec:eval:compare}. Finally, we evaluate the achieved swarm performance in a large-scale, randomized simulation study as described in \cref{sec:eval:perf}.


\subsection{Methodology}
\label{sec:eval:methodology}
We use two simulation environments. The first is a kinematic simulation, which is programmed in Python to be fast and scalable. It therefore allows us to analyze numerous test settings, resulting in increased statistical significance. 
The second simulation is physics-based, intended to verify the realism of the kinematic approximations. It is implemented in ROS2 using  Gazebo Harmonic. In the next subsection, the scenario generation is described, followed by a description of the metrics employed in the analysis.

\subsubsection{Scenario Generation}

\begin{figure}[hbt]
    \centering
    \begin{tikzpicture}
        \begin{axis}[
            axis lines = left,
            xlabel={Trailer count, $N_i$ [1]},
            ylabel={Probability},
            domain=0:10,
            samples=100,
            ymin=0,
            ymax=0.21,
            ytick={0,0.1,0.2},
            height=.25\linewidth,
            width=.45\linewidth
        ]
            \addplot [
                thick, blue
            ]
            {(x/(3^2)) * exp((-x^2)/(2*3^2))};
        \end{axis}
    \end{tikzpicture}
    \hfill
    \begin{tikzpicture}
        \begin{axis}[
            axis lines = left,
            xlabel={Truck length, $l_i^0$ [\si{\meter}]},
            ylabel={Probability},
            domain=2:12.05,
            samples=100,
            ymin=0,
            ymax=0.35,
            height=.25\linewidth,
            width=.45\linewidth
        ]
            \addplot [
                thick, red
            ]
            {0.5 * 1/(sqrt(2*pi*0.6^2)) * exp(-(x-4)^2/(2*0.6^2)) +
             0.5 * 1/(sqrt(2*pi*1.2^2)) * exp(-(x-10.7)^2/(2*1.2^2))};
            \draw[thick, red] (axis cs: 12, 0.093) -- (axis cs: 12, 0); 
        \end{axis}
    \end{tikzpicture}
    \caption{Random HAV generation: distribution of trailer count and of truck length.}
    \label{fig:hav-gen}
\end{figure}
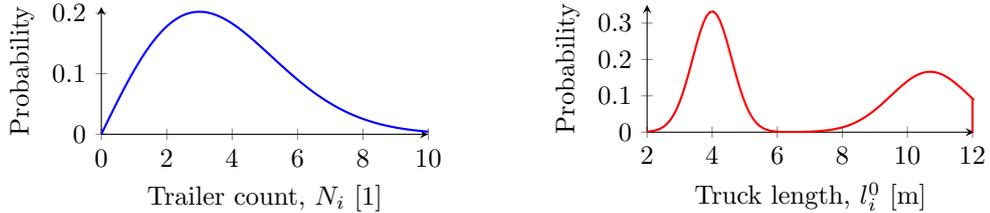

To model real-world truck-trailer combinations, each HAV $i$ is composed of a single truck and $N_i$ trailers, where $N_i\in\{1,\dots,10\}$ is sampled from a Rayleigh distribution with $\sigma=3$ (see \cref{fig:hav-gen}). The truck length $l_i^0\in[2\si{\meter},12\si{\meter})$ is sampled from a mixed Gaussian distribution with $\mu_1=4\si{\meter},\sigma_1=0.6,\mu_2=10.7\si{\meter},\sigma_2=1.2$ (see \cref{fig:hav-gen}), as most trucks are either short semi-trucks or long cargo trucks. Each trailer length $l_i^j\in[2\si{\meter},12\si{\meter})$ is independently sampled from a uniform distribution.

For the kinematic simulation, scenarios are randomized following \cref{sec:problem:scenario}. Torus dimensions are determined by collision density $\rho$, defined as the percentage of space occupied by HAVs' circular representations. The torus edge length $d_\text{torus}$ satisfies $d_\text{torus}^2 = \frac{1}{\rho} \sum_{i=0}^{N_H}(\pi d_i^2)$ for swarm size $N_H$ and HAV collision radii $d_i$. When all HAVs reach their first goal pose, a final second goal is assigned to all HAVs. 
Simulations terminate when all HAVs reach both consecutive goals, a deadlock is detected, or the maximum time-step limit is exceeded. Simulation-code is available at Zenodo \cite{zenodoANTS2025}.

The physics-based Gazebo simulation is restricted to a limited number of vehicle configurations due to the increased setup complexity. In this paper, we focus on airport baggage transport HAVs, as they are a common example of vehicles with multiple passive trailers. We model versions with two and four baggage carts. Note that one cart consists of two trailers, as its front axle is connected to the body through a passive yaw joint, therefore yielding four and eight trailers, respectively. Due to the simulation constraints, we cannot employ torus worlds here and revert to hand-crafted scenarios of three and ten HAVs.   

\subsubsection{Metrics}
\label{sec:eval:metrics}

We distinguish between two \textit{failure} modes: a \textit{deadlocked simulation} occurs when no HAV is moving, and every HAV has either arrived at its goal or is obstructed by its block mask. A \textit{livelocked simulation} is identified when the maximum iteration count is exceeded; this detection method may yield false positives and capture \textit{partial deadlocks}, i.e., one HAV is deadlocked while another remains livelocked.

Both deadlock and livelock occurrences are reported at two granularity levels. At the global level, we quantify the percentage of experimental runs exhibiting the respective failure mode. At the HAV level, we report the proportion of individual HAVs that failed to reach their goals within deadlocked or livelocked simulations, respectively.

Two additional metrics characterize operational behavior. \textit{Average speed} denotes the ratio of traveled distance to elapsed time per HAV. The time when HAVs are waiting in their goal position is not counted. Note that when multiple actions yield identical interest values, the algorithm prefers higher-velocity actions. Therefore, reduced speeds occur exclusively when HAVs encounter path blockages imposed by Jackknife Prevention, Collision Prevention, or Evade Attraction mechanisms. \textit{Path deviation} represents the ratio of actual traveled distance to the initial planned, minimal path length per HAV. A high average speed combined with low path deviation is therefore a sign of high throughput, and a short task makespan.

When aggregating across multiple simulation runs, we employ macro-averag\-ing: metrics are computed individually per run, and the mean value across all runs is reported.

\subsection{Parameter Study}
\label{sec:eval:params}

Prevent-Jack behaviors incorporate parameters that influence algorithmic performance. While some parameters can be determined analytically, others map to human-like characteristics such as daringness, impatience, or risk tolerance. We tune these parameters using the kinematic simulation as follows:

\subsubsection{Design}
\label{sec:eval:params:design}

We begin with the parameterization defined in \cref{sec:prevent-jack}, which was tuned by preliminary experiments. We will refer to it as the \textit{baseline} and test its quality, assuming that parameter effects are largely independent, especially from one behavior to the other. We vary one parameter at a time from the baseline and evaluate multiple values. For each parameter, we select the value that minimizes failures; in case of ties, we prefer fewer deadlocks. The parameters and candidate values are listed in \cref{fig:param-study}.

Additionally, we analyze the effect of the resolution of the context map. To ensure full coverage of the action space, we select values for the context-map actions $(v,\phi)$ from the sets $v \in V_{n_v}$ and $\phi \in \Phi_{n_\phi}$, which are evenly spaced between their respective bounds $V_{n_v} \subset [\SI{0}{\meter\per\second}, \SI{4}{\meter\per\second}], \Phi_{n_\phi} \subset [\SI{-50}{\degree}, \SI{50}{\degree}]$. The evaluated resolutions are $n_v \in \{2,3,5,7,9\}$ and $n_\phi \in \{3, 5,7,9\}$, guaranteeing that $\SI{0}{\degree}\in\Phi_{n_\phi}$. As interpolation, we use cubic interpolation if $n_v \ge 4$ and $n_\phi \ge 4$ otherwise we revert to linear interpolation, since cubic interpolation is preferable for its smoothness but not reliably applicable when either axis has fewer than four support points.

We then aggregate the per-parameter optima into a jointly tuned \textit{final} configuration and evaluate its performance. Each parameter set is tested on an identical set of 100 randomly generated scenarios to ensure comparability. Each scenario features five HAVs with a randomly drawn trailer count $\in \{1,\dots,10\}$ (cf. \cite{schonnagelAVOIDJACKAvoidanceJackknifing2025}), and a collision density of $\rho=\SI{12}{\percent}$. Simulations are capped at 10{,}000 time steps, with a total of 3{,}700 runs.
The optimized parameter set will be used for the remaining simulations.

\subsubsection{Results}
\label{sec:eval:params:res}

\begin{figure}[hbt]
    \centering
    \includegraphics{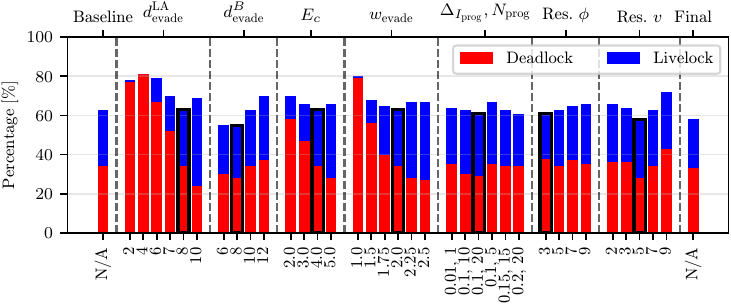}
    \caption{Deadlock and livelock percentages observed in the parameter study. The top x-axis shows the parameter, while the bottom x-axis marks the tested values. Outlined bars mark selected values. Res. refers to the context map resolution.}
    \label{fig:param-study}
\end{figure}

The results of the parameter study are shown in \cref{fig:param-study}. 
For the Evade Attraction lookahead distance $d_\text{evade}^\text{LA}$, its exponent $E_c$, and its weight $w_\text{evade}$, we observe a shift from deadlocks to livelocks as parameter values increase. Low $d_\text{evade}^\text{LA}$ and $w_\text{evade}$ markedly increased deadlocks, consistent with our preliminary experiments and underscoring the importance of the Evade Attraction behavior. Both parameters are positively correlated with path deviation, while average speed remains largely unchanged.

Regarding context-map resolution for velocities $v$, the median resolution appears preferable. For steering angles $\phi$, lower resolution yields fewer failures, accompanied by slightly higher average speed and path deviation. For progress attraction, a slower increment appears to be preferable. 

The final parameter set yields fewer failures than the baseline, although some alternative combinations achieve comparable performance (see the Evade Attraction threshold $d_\text{evade}^\text{B}$). This suggests non-independent effects among parameters, being trapped in a local minimum, or a well-chosen baseline. A comprehensive analysis of all combinations is beyond the scope of this paper; we proceed with the improved parameters.

\subsection{Physics-Based vs. Kinematic Simulation}
\label{sec:eval:gazebo}

While the kinematic simulation is much faster and more compute-efficient, therefore allowing for more runs to be executed, it is crucial to validate its realism. For this purpose, we utilize a second, physics-based simulation as follows: 

\subsubsection{Design}
\label{sec:eval:gazebo:draft}

We create two scenarios to be executed in both simulations, with three and ten airport baggage HAVs, respectively. Both simulations use standard worlds (non-torus) and are set to a control rate of \SI{20}{\hertz}. We use ground truth localization, position communication, and articulation angle reading for the robots, with an update rate of \SI{50}{\hertz} in the physics-based simulation and immediate updates in the kinematic simulations. The physics-based simulation also models limited longitudinal and steering accelerations and velocities, while the kinematic simulation assumes unbounded accelerations and unbounded steering velocity. Furthermore, the physics-based simulation also considers more dynamic effects, such as wheel slippage and grip. 

We then compare the driven paths of all HAVs over time.

\subsubsection{Results}
\label{sec:eval:gazebo:res}

\Cref{fig:gazebo_comp} shows the physics-based and kinematic simulations side by side. For the experiment with three HAVs, we see almost no visual difference in driven paths. When scaling to ten HAVs, we observe some variation, depending on the configured acceleration limits. The higher the allowed acceleration, the more similar the driven paths become. When restricting acceleration, we observe some HAVs (e.g., number 5) arriving with some small delay at a rendezvous point, which leads to it choosing a different evasive action. This deviation then propagates further through the experiment. 

\begin{video}[hbt]
    \centering
    \includegraphics[width=\linewidth]{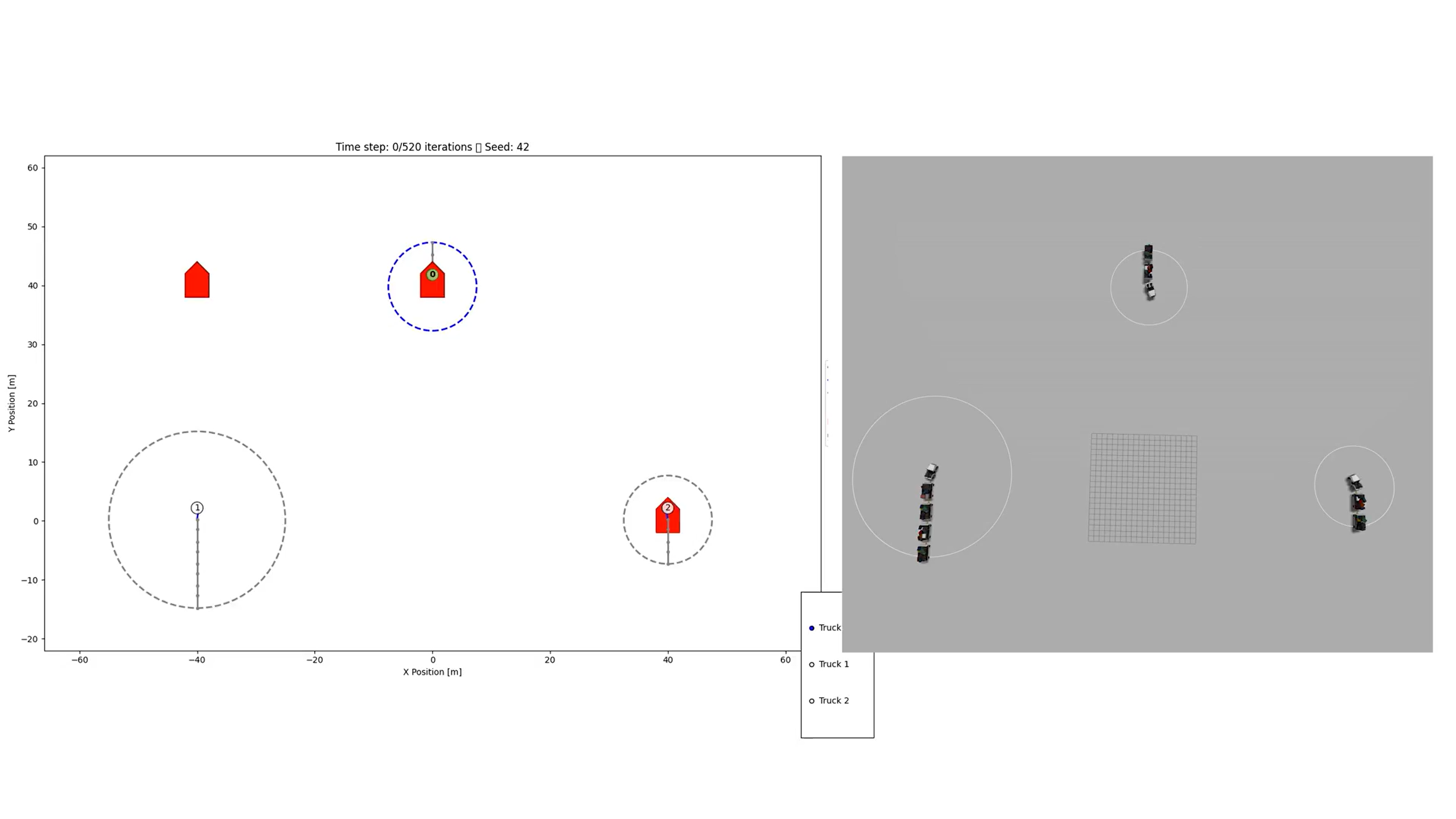}
    \caption{Comparison of physics-based and kinematic simulations. In the PDF version of this article, please refer to \textbf{vid3\_sim\_comparision.mp4} in the accompanying archive to view the video.}
    \label{fig:gazebo_comp}
\end{video}

While these deviations highlight the simplified nature of the kinematic simulation, they also showcase the adaptability of the algorithm to unforeseen events, such as delays in execution, which is the main goal of this work. 
Further, we observe large differences in computational demand. While the physics-based simulation struggles to maintain $0.04\times$ real-time factor and requires a GPU, the kinematic simulation runs at $1.95\times$ real-time factor on a single CPU core, allowing for parallelization. Therefore, we will use the kinematic simulation for the large-scale simulation study in \cref{sec:eval:perf} with randomization of the vehicle configurations, start poses, and goal poses to cover a large variety of possible evasive decisions. 

\subsection{Comparison to Baselines}
\label{sec:eval:compare}

As described in \cref{sec:sota}, there is a lack of established baselines for the (currently niche) problem addressed in this work. A comparison to systems with global pre-planning and complete future knowledge would primarily reflect differences in information structure rather than the quality of the reactive coordination strategy itself. Decentralized methods, such as ORCA, are not yet extended to incorporate HAVs. Consequently, we restrict our quantitative evaluation to the most closely related method, our simple Avoid-Jack \cite{schonnagelAVOIDJACKAvoidanceJackknifing2025}.

\subsubsection{Design}
\label{sec:eval:compare:design}

For fairness, we use the kinematic simulation setup described earlier and match the configuration of the Avoid-Jack study as closely as possible. The only change in world representation is that Avoid-Jack was evaluated in effectively infinite worlds, while Prevent-Jack operates in torus worlds; apart from this change in topology, the scenarios and parameters follow the same randomization. In addition, Prevent-Jack enforces non-overlap of circular HAV approximations, whereas Avoid-Jack only soft-constrains the same, which effectively provides Prevent-Jack with less usable space.

For Prevent-Jack, we perform 2{,}500 randomized experiments each for swarms of size one and two. We do not evaluate larger swarms in this direct comparison, as two HAVs was the upper limit in the original Avoid-Jack study, which used 4{,}500 experiments per swarm size. We compare the observed jackknifing rate, collision rate, and \textit{task-completion} rate, defined as the fraction of experiments in which all HAVs reach both consecutive goals without entering deadlock or livelock. Our hypotheses are that Prevent-Jack: (i) exhibits no jackknifing, (ii) for a swarm size of one, achieves deadlock- and livelock-free task completion, and (iii) for a swarm size of two, prevents collisions while increasing the task-completion rate relative to Avoid-Jack.

\subsubsection{Results}
\label{sec:eval:compare:res}

As summarized in Table~\ref{tab:comp-avoid-jack}, Prevent-Jack consistently outperforms our prior Avoid-Jack baseline for swarms of size one and two, supporting all three hypotheses from \cref{sec:eval:compare:design}. Across both swarm sizes, jackknifing is eliminated, with rates reduced from \SI{0.16}{\percent} and \SI{1.1}{\percent} to \SI{0.0}{\percent}, confirming hypothesis (i) and demonstrating the effectiveness of the Jackknife Prevention Behavior within the Context Steering framework.
\begin{table}[hbt]
    \centering
    \caption{Comparison to Avoid-Jack. Task completion: HAVs reached both consecutive goals, i.e., no deadlock or livelock.}
    \label{tab:comp-avoid-jack}
    \begin{tabular}{lcccc}
        \toprule
        \multirow{2}{*}{Metric} & \multicolumn{2}{c}{Swarm Size $N_H=1$} & \multicolumn{2}{c}{Swarm Size $N_H=2$} \\
        \cmidrule(lr){2-3}\cmidrule(lr){4-5}
         & Avoid-Jack & Prevent-Jack & Avoid-Jack & Prevent-Jack \\
        \midrule
        Task Completion & \SI{83.4}{\percent} & \SI{100.0}{\percent} & \SI{65.1}{\percent} & \SI{73.2}{\percent} \\
        Jackknifing   & \SI{0.16}{\percent} & \SI{0.0}{\percent} & \SI{1.1}{\percent} & \SI{0.0}{\percent} \\
        Collisions    & -- & -- & \SI{0.3}{\percent} & \SI{0.0}{\percent} \\
        \bottomrule
    \end{tabular}
\end{table}

For swarms of size one, Prevent-Jack increases task completion from \SI{83.4}{\percent} to \SI{100.0}{\percent}, i.e., in every run the HAV reaches both consecutive goals without dead- or livelock, confirming hypothesis (ii). This improvement is most likely attributable to the path-following controller described in \cref{sec:behavior:dubins}.

For swarms of size two, Prevent-Jack simultaneously improves safety and task performance: task completion increases from \SI{65.1}{\percent} to \SI{73.2}{\percent}, while collisions drop from \SI{0.3}{\percent} to \SI{0}{\percent}. This confirms hypothesis (iii) and can be attributed to the Collision Prevention Behavior integrated via context steering. Together with the absence of jackknifing, these results are consistent with the algorithm’s design for scalable, jackknife- and collision-free operation in denser traffic, supporting its application to larger swarms, as examined in the next subsection.

Further, Avoid-Jack is limited to a maximum of two vehicles, whereas Prevent-Jack scales to larger fleets.

\subsection{Swarm Performance}
\label{sec:eval:perf}

Finally, we evaluate the performance of the HAV swarm with the proposed method. We utilize the kinematic simulation for its increased scalability, allowing for large-scale simulations as follows:

\subsubsection{Design}
\label{sec:eval:perf:design}

We evaluate algorithm performance across varying swarm sizes $N_H\in\{2, 3, 5, 7, 10, 20\}$ and collision densities $\rho \in \{\SI{5}{\percent}, \SI{10}{\percent}, \SI{15}{\percent}, \SI{20}{\percent}, \SI{25}{\percent}\}$. Each combination is tested on 500 independently generated scenarios, as scenario generation depends on both factors. To account for increased computational complexity arising from larger swarm sizes and higher densities, we raise the time-step limit to $20{,}000$ iterations, thereby reducing false positive livelock detections. This experimental design yields a total of $15{,}000$ simulation runs.

\subsubsection{Results}
\label{sec:eval:perf:res}

\begin{figure}[hbt]
    \centering
    \includegraphics{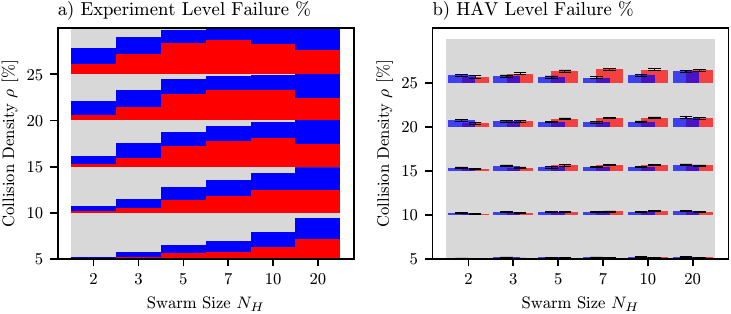}
    \caption{Swarm Performance over $\rho$ and $N_H$: Each cell shows distribution of successful (gray), deadlocked (red), and livelocked (blue) experiments. \textbf{a}: Overall experiment results; \textbf{b}: Average percentage of affected HAVs with \SI{95}{\percent} confidence intervals.}
    \label{fig:swarm_perf_deadlocks}
\end{figure}

Percentages of deadlocked and livelocked runs are shown on the left of \cref{fig:swarm_perf_deadlocks} across all combinations of swarm size $N_H$ and collision density $\rho$. Both failure types increase with $N_H$ and $\rho$. Beyond the threshold where no runs succeed, further increases in $N_H$ partially shifts failures from dead- to livelock. Raising the time-step limit from 10,000 to 20,000 iterations decreased detected livelocks by a maximum of \SI{3.7}{\percent} (\SI{54.7}{\percent} to \SI{50.9}{\percent}), with most improvements in the \SI{1}{\percent}-\SI{2}{\percent} range. This suggests that few false-positive livelock detections remain, though the metric may still capture partial deadlocks.

A finer-grained view is provided on the right of Fig. \ref{fig:swarm_perf_deadlocks}, which reports the fraction of HAVs affected within 
runs. 
The HAV-level affect rates also rise with $N_H$ and $\rho$, but more gradually than the experiment-level rates. Even in the most challenging configuration, on average \SI{73}{\percent}/\SI{69}{\percent} of the HAVs still reach their goals.

\begin{figure}[hbt]
    \centering
    \includegraphics{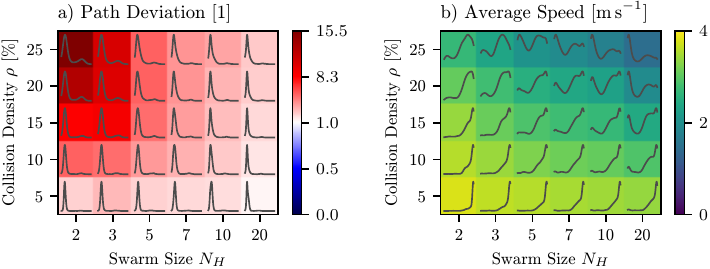}
    \caption{Swarm performance over $\rho$ and $N_H$. Cell colors represent the mean values, and gray lines show KDEs of respective data. \textbf{a}: Path deviation $[1]$ with KDEs in $\log$-scale; \textbf{b}: Average speed $[\si{\meter\per\second}]$.}
    \label{fig:swarm_perf:speed_and_deviation}
\end{figure}

\cref{fig:swarm_perf:speed_and_deviation} presents heatmaps for path deviation and average speed. Cell color encodes the mean value, and overlaid curves depict Kernel Density Estimates (KDE) \cite{rosenblattKDE1956,parzenKDE1962} of the corresponding distributions. Mean path deviation (left) decreases with increasing swarm size and decreasing collision density. No cell has a mean below 1, implying that unfinished paths do not constitute a majority. Differences in the KDEs are only apparent on a log scale: all cells exhibit a strong peak at low deviation, and higher means are associated with an additional smaller peak at larger deviations, going up to $500\times$, most probably relating to livelocked cases.

Mean average speed (\cref{fig:swarm_perf:speed_and_deviation} right) decreases with increasing swarm size and collision density. For higher mean speeds, KDEs are skewed toward large speeds; as the mean decreases, KDEs become bimodal with mass at both low and high speeds. At the largest $N_H$ and highest $\rho$, the distribution concentrates at low speeds. This suggests a polarization wherein HAVs move either near maximum speed or very slowly. A plausible explanation is undetected partial deadlocks, where some HAVs remain stationary while others continue moving—an occurrence that becomes more likely as the swarm grows.

Considering path deviation and average speed jointly reveals a swarm-size–dependent trade-off. Larger swarms move more slowly with lower deviation, consistent with increased waiting, whereas smaller swarms move faster with greater deviation, consistent with more evasion. While this effect matches intuition, it may be influenced by how the toroidal world scales with swarm size, implying that a fixed collision density does not perfectly equalize scenario difficulty. As expected, higher collision density induces both more evasion and more waiting.

\begin{video}[hbt]
    \centering
    \includegraphics[width=1.0\linewidth]{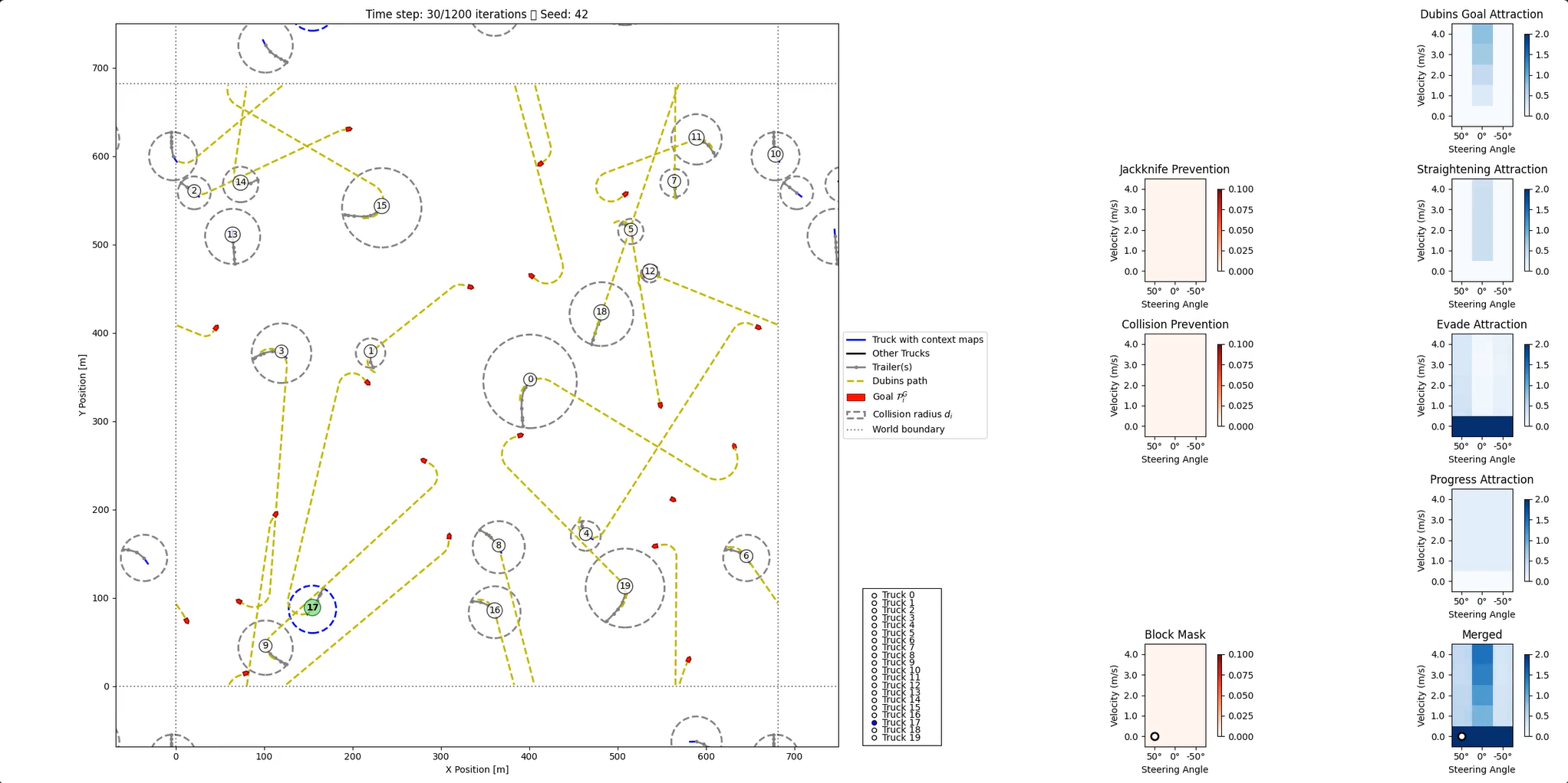}
    \caption{Exemplary simulation results, including livelocks and partial deadlocks. In the PDF version of this article, please refer to \textbf{vid4\_exp\_results.mp4} in the accompanying archive to view the video.}
    \label{fig:sim-results}
\end{video}

Exemplary runs are shown in \cref{fig:sim-results} including livelocks and (partial) deadlocks.
Finally, no jackknifing events or collisions were observed in the simulations, corroborating that Prevent-Jack effectively prevents these.


\section{Conclusion and Future Work}
\label{sec:clusion}

This work extends our prior work on decentralized HAV fleets by presenting the first algorithm that guarantees prevention of jackknifing and mutual collisions in complex, randomized multi-HAV scenarios, tested with up to ten trailers per vehicle. To the best of our knowledge, it is the first purely reaction-based algorithm for large, decentralized HAV fleets and one of the first works to implement highly kinematically constrained vehicles in swarm robotics. Further, we showcased the advantages of multi-objective behavior merging by transferring the context steering framework to robotics. An application to other fields and use cases might be of interest. Despite this advance, practical deployment of our proposed approach is constrained by deadlock and livelock occurrences, which require systematic analysis and mitigation.

Future work might address these limitations through several approaches. Deadlock mitigation may include movement prediction (similar to MPC, ORCA), obstacle-aware path planning that temporarily models other HAVs as static obstacles, right-of-way scheduling or negotiation, or enabling temporary goal abandonment to facilitate passage. Additionally, the approach could be extended to bounded, cluttered environments through reactive or pre-planning strategies, and to limited intervehicle communication rate or bandwidth, with real-world robot experiments validating practical feasibility.

Algorithm refinement through hyperparameter optimization, e.g., using evolutionary multi-objective optimization (EMO), could enhance performance, complemented by partial deadlock detection capabilities for improved analysis. 
Further, alternative approaches, such as ORCA~\cite{vandenBergORCA11}, rolling horizon evolutionary algorithms~\cite{perez13RHEA}, or Monte Carlo Tree Search~\cite{bone23MonteCarlo} might be extended to HAVs and compared to the approach presented in this work. 

Most critically, however, future work should investigate more space-efficient footprint approximations for mutual collision avoidance, which would simultaneously address deadlock prevention while improving overall system efficiency.
Then finally, application and possibly adaptation of the algorithm to specific use cases and dynamic vehicle constraints, such as limited accelerations and movement smoothness, might be evaluated.


\backmatter

\bmhead{Acknowledgements}

This research was funded by EFRE Saxony-Anhalt, grant number ZS/2023/12/182177, and supported by the Fraunhofer Internal Programs under Grant No. Attract 40-11882. Further, we would like to thank Mr. Maximilian Stahr of Fraunhofer IVI for editing the videos.


\section*{List of Symbols}

\subsection*{Indices and Swarm}

\begin{tabularx}{\linewidth}{>{\raggedright}p{.15\linewidth} X l}
\toprule
\textbf{Symbol} & \textbf{Description} & \textbf{Unit} \\
\midrule
$N_H$    & Swarm Size, i.e., Total number of HAVs in the swarm & --- \\
$i,\, h$ & Indices of individual HAVs, $i,h \in \{0,\ldots,N_H-1\}$ & --- \\
$N_i$    & Number of trailers of HAV $i$ & --- \\
$j$      & Trailer index, $j \in [1, N_i]$ & --- \\
$k$      & Virtual axle index & --- \\
\bottomrule
\end{tabularx}

\subsection*{Vehicle Modeling}

\begin{tabularx}{\linewidth}{>{\raggedright}p{.15\linewidth} X l}
\toprule
\textbf{Symbol} & \textbf{Description} & \textbf{Unit} \\
\midrule
$(x_i^k,\, y_i^k)$  & Position of the $k$-th virtual axle of HAV $i$ & m \\
$\Theta_i^0$        & Heading of the truck of HAV $i$ & rad \\
$\Theta_i^j$        & Heading of trailer $j$ of HAV $i$ & rad \\
$\delta_i^j$        & Articulation angle between trailer $j$ and its predecessor,
                      $\Theta_i^j - \Theta_i^{j-1}$ & rad \\
$l_i^0$             & Wheelbase of the truck of HAV $i$ & m \\
$l_i^j$             & Wheelbase of trailer $j$ of HAV $i$ & m \\
\midrule
$v_i^0$            & Longitudinal speed of the truck of HAV $i$ & m\,s$^{-1}$ \\
$v_i^j$            & Propagated speed of trailer $j$ of HAV $i$ & m\,s$^{-1}$ \\
$\phi_i$           & Steering angle of HAV $i$ & rad \\
$\phi_i^{\max}$    & Maximum steering angle of HAV $i$ (symmetric) & rad \\
$v_i^\text{min},v_i^\text{max}$ & Minimum/maximum velocity of HAV $i$ & \si{\meter\per\second} \\
\midrule
$R_i^{\min}$        & Minimum stable turning radius of HAV $i$ & m \\
$\Delta t$         & Simulation time step & s \\
\midrule
$d_i$           & Circular collision radius (footprint) of HAV $i$ & m \\
$d_{i,h}$       & Combined collision distance between HAVs $i$ and $h$ & m \\
$\mathcal{P}_i$ & Polygonal chain of HAV $i$ connecting successive axles & --- \\
\bottomrule
\end{tabularx}

\subsection*{Poses and Scenario}

\begin{tabularx}{\linewidth}{>{\raggedright}p{.15\linewidth} X l}
\toprule
\textbf{Symbol} & \textbf{Description} & \textbf{Unit} \\
\midrule
$P_i^S$              & Start pose of HAV $i$,\;
                       $\{p_{x,i}^S,\, p_{y,i}^S,\, p_{\Theta,i}^S\}$ & --- \\
$P_i^G$              & Goal pose of HAV $i$,\;
                       $\{p_{x,i}^G,\, p_{y,i}^G,\, p_{\Theta,i}^G\}$ & --- \\
$\rho$               & Collision density (fraction of space occupied
                       by HAV footprints) & \% \\
$d_{\mathrm{torus}}$ & Edge length of the toroidal simulation world & m \\
\bottomrule
\end{tabularx}

\subsection*{Context Steering}

\begin{tabularx}{\linewidth}{>{\raggedright}p{.15\linewidth} X l}
\toprule
\textbf{Symbol} & \textbf{Description} & \textbf{Unit} \\
\midrule
$C_I^k$      & Interest context map of behavior $k$ & --- \\
$w_k$        & Weight of interest context map $k$, default$ = 1$ & --- \\
$C_D^k$      & Danger context map of behavior $k$ & --- \\
$C_I^S$      & Merged interest map (weighted sum) & --- \\
$C_I^F$      & Filtered interest map (after block mask applied) & --- \\
$B$          & Binary block mask (prohibited actions) & --- \\
$\epsilon_D$ & Danger threshold for block mask construction & --- \\
\midrule
$n_v$        & Context map velocity resolution & --- \\
$n_\phi$     & Context map steering angle resolution & --- \\
$V_{n_v}$    & Set of discrete velocities for context map actions & --- \\
$V_{n_\phi}$ & Set of discrete steering angles for context map actions & --- \\
$n_v^\text{int}$        & Context map velocity interpolation resolution & --- \\
$n_\phi^\text{int}$     & Context map steering angle interpolation resolution & --- \\

\bottomrule
\end{tabularx}

\subsection*{Path Following Controller}

\begin{tabularx}{\linewidth}{>{\raggedright}p{.15\linewidth} X l}
\toprule
\textbf{Symbol} & \textbf{Description} & \textbf{Unit} \\
\midrule
$e_P^{\max}$                    & Maximum tolerated path deviation
                                  before replanning & m \\
$p_C$                           & Nearest path sample point                                                  $(x_C,\,y_C)^\top$ to the
                                  truck's rear axle & m \\
$p_P$                           & Lookahead point $(x_P,\,y_P)^\top$ on                                      the path & m \\
$\Theta_P$                      & Heading of the path at the lookahead                                       point & rad \\
$l_C$                           & Lookahead distance,
                                  $l_C = f_C \cdot l_i^0$ & m \\
$f_C$                           & Lookahead factor & --- \\
\midrule
$e_H$                           & Heading error,
                                  $\Theta_P - \Theta_i^0$ & rad \\
$e_P$                           & Cross-track error (Euclidean distance
                                  to $p_C$) & m \\
$k_e$                           & Cross-track error gain
                                  (Stanley controller) & --- \\
$\phi_i^P$                   & Pure-pursuit steering component & rad \\
$\phi_i^S$                   & Stanley steering component & rad \\
$\phi_i^C$                   & Combined path-following steering
                                  angle & rad \\
$\sigma_\phi$                & Standard deviation of goal-attraction
                                  Gaussian in steering dimension & rad \\
$\sigma_v$                      & Standard deviation of goal-attraction
                                  Gaussian in velocity dimension
                                & m\,s$^{-1}$ \\
\bottomrule
\end{tabularx}

\subsection*{Straightening Attraction Behavior}

\begin{tabularx}{\linewidth}{>{\raggedright}p{.15\linewidth} X l}
\toprule
\textbf{Symbol} & \textbf{Description} & \textbf{Unit} \\
\midrule
$I_J$                & Straightening attraction interest value & --- \\
$f_{IJ}(\delta_i^j)$ & Articulation-dependent interest function
                       per hitch joint & --- \\
\bottomrule
\end{tabularx}

\subsection*{Collision Behaviors}

\begin{tabularx}{\linewidth}{>{\raggedright}p{.15\linewidth} X l}
\toprule
\textbf{Symbol} & \textbf{Description} & \textbf{Unit} \\
\midrule
$g_h$                              & Separation distance (gap) from HAV $i$'s simulated endpoint
                                     to HAV $h$ & m \\
$d_\text{coll}^\text{LA}$          & Lookahead distance for Collision
                                     Prevention danger behavior & m \\
$d_\text{evade}^\text{LA}$         & Lookahead distance for Evade
                                     Attraction interest behavior & m \\
$d_\text{evade}^B$                 & Distance threshold beyond which
                                     no evasion penalty applies & m \\
$E_c$                              & Exponent controlling steepness of
                                     evasion interest reduction & --- \\
$p(g_h)$                           & Distance-dependent interest penalty
                                     for neighboring HAV $h$ & --- \\
$w_\text{evade}$                   & Weight of Evade Attraction behavior
                                     in interest merging & --- \\
$I_\text{evade}$                   & Evade Attraction interest value (of an action) & --- \\
\bottomrule
\end{tabularx}

\subsection*{Progress Attraction Behavior}

\begin{tabularx}{\linewidth}{>{\raggedright}p{.15\linewidth} X l}
\toprule
\textbf{Symbol} & \textbf{Description} & \textbf{Unit} \\
\midrule
$n_\text{standstill}$      & Counter of consecutive standstill
                             time steps & --- \\
$N_\text{prog}$            & Period between progress interest
                             increments & steps \\
$\Delta_{I_\text{prog}}$   & Progress Attraction interest increment magnitude per period & --- \\
$I_\text{prog}$            & Progress Attraction interest value & --- \\
\bottomrule
\end{tabularx}


\section*{Declarations}

\subsection*{Funding}
This research was funded by EFRE Saxony-Anhalt under grant ZS/2023/12/182177, supporting A.S. and M.D., and by the Fraunhofer Internal Programs under grant Attract 40-11882, supporting C.S.

\subsection*{Competing interests}
The authors declare no competing interests.

\subsection*{Ethics approval and consent to participate}
This study did not involve human participants, human data, or animals, and therefore did not require ethics approval or consent to participate.

\subsection*{Consent for publication}
Not applicable.

\subsection*{Data availability}
The data generated and analyzed during this study are reproducible using the Zenodo archive associated with this article \cite{zenodoANTS2025}, which contains the scripts used to run the kinematic simulations, perform the analyses, and generate all plots and animations. The resulting datasets underlying the figures and quantitative results are also available from the corresponding author upon reasonable request. Visualizations of context steering and simulated experiments are provided in the accompanying videos.

\subsection*{Materials availability}
Not applicable. No new hardware or physical materials were generated in this study.

\subsection*{Code availability}
The code used to run the kinematic simulations, perform the analyses, and generate the plots and animations is available via the Zenodo repository associated with this work \cite{zenodoANTS2025} under a Creative Commons Attribution–NonCommercial–ShareAlike license and is intended for academic, non-commercial use; the authors do not support military applications of this code. Code for the Gazebo-based simulations is not publicly archived but is available from the corresponding author upon reasonable request; third-party 3D models used in these simulations are not redistributed, as they require separate licenses from their respective rights holders.

\subsection*{Author contribution}
A.B. broached the idea, conceived the study, developed the algorithm and behaviors, implemented the simulations, and wrote the manuscript. M.D. and C.S. equally contributed to the study design through regular discussions during development and to manuscript revisions. S.M. initiated the idea, supervised the project, and contributed to manuscript revision.


\bibliography{myBib}

@article{dubinsCurvesMinimalLength1957,
  title = {On {{Curves}} of {{Minimal Length}} with a {{Constraint}} on {{Average Curvature}}, and with {{Prescribed Initial}} and {{Terminal Positions}} and {{Tangents}}},
  author = {Dubins, L. E.},
  year = {1957},
  journal = {American Journal of Mathematics},
  volume = {79},
  number = {3},
  eprint = {2372560},
  eprinttype = {jstor},
  pages = {497--516},
  publisher = {Johns Hopkins University Press},
  issn = {0002-9327},
  url = {https://doi.org/10.2307/2372560},
  urldate = {2022-10-24}
}

@article{kantEfficientTrajectoryPlanning1986,
  title = {Toward {{Efficient Trajectory Planning}}: {{The Path-Velocity Decomposition}}},
  shorttitle = {Toward {{Efficient Trajectory Planning}}},
  author = {Kant, Kamal and Zucker, Steven W.},
  year = {1986},
  month = sep,
  journal = {The International Journal of Robotics Research},
  volume = {5},
  number = {3},
  pages = {72--89},
  publisher = {SAGE Publications Ltd STM},
  issn = {0278-3649},
  url = {https://doi.org/10.1177/027836498600500304},
  langid = {english}
}

@inproceedings{kepplerPrioritizedMultiRobotVelocity2020,
  title = {Prioritized {{Multi-Robot Velocity Planning}} for {{Trajectory Coordination}} of {{Arbitrarily Complex Vehicle Structures}}},
  booktitle = {2020 {{IEEE}}/{{SICE International Symposium}} on {{System Integration}} ({{SII}})},
  author = {Keppler, F. and Wagner, S.},
  year = {2020},
  editor = {IEEE},
  month = jan,
  pages = {1075--1080},
  issn = {2474-2325},
  url = {https://doi.org/10.1109/SII46433.2020.9026256}
}

@article{kumarLyapunovBasedControlSwarm2015,
  title = {Lyapunov-{{Based Control}} for a {{Swarm}} of {{Planar Nonholonomic Vehicles}}},
  author = {Kumar, Sandeep Ameet and Vanualailai, Jito and Sharma, Bibhya},
  year = {2015},
  month = dec,
  journal = {Math.Comput.Sci.},
  volume = {9},
  number = {4},
  pages = {461--475},
  issn = {1661-8289},
  url = {https://doi.org/10.1007/s11786-015-0243-z},
  urldate = {2025-03-14},
  langid = {english}
}

@article{diasSwarmRoboticsPerspective2021,
  title = {Swarm {{Robotics}}: {{A Perspective}} on the {{Latest Reviewed Concepts}} and {{Applications}}},
  shorttitle = {Swarm {{Robotics}}},
  author = {Dias, Pollyanna G. Faria and Silva, Mateus C. and Rocha Filho, Geraldo P. and Vargas, Patr{\'i}cia A. and Cota, Luciano P. and Pessin, Gustavo},
  year = {2021},
  month = jan,
  journal = {Sensors},
  volume = {21},
  number = {6},
  pages = {2062},
  publisher = {Multidisciplinary Digital Publishing Institute},
  issn = {1424-8220},
  url = {https://doi.org/10.3390/s21062062},
  urldate = {2025-04-03},
  copyright = {http://creativecommons.org/licenses/by/3.0/},
  langid = {english}
}

@inproceedings{hauertReynoldsFlockingReality2011,
  title = {Reynolds Flocking in Reality with Fixed-Wing Robots: {{Communication}} Range vs. Maximum Turning Rate},
  shorttitle = {Reynolds Flocking in Reality with Fixed-Wing Robots},
  booktitle = {2011 {{IEEE}}/{{RSJ International Conference}} on {{Intelligent Robots}} and {{Systems}}},
  author = {Hauert, Sabine and Leven, Severin and Varga, Maja and Ruini, Fabio and Cangelosi, Angelo and Zufferey, Jean-Christophe and Floreano, Dario},
  year = {2011},
  editor = {IEEE},
  month = sep,
  pages = {5015--5020},
  issn = {2153-0866},
  url = {https://doi.org/10.1109/IROS.2011.6095129},
  urldate = {2025-04-08}
}

@inproceedings{sordalenConversionKinematicsCar1993,
  title = {Conversion of the Kinematics of a Car with n Trailers into a Chained Form},
  booktitle = {[1993] {{Proceedings IEEE International Conference}} on {{Robotics}} and {{Automation}}},
  author = {Sordalen, O.J.},
  year = {1993},
  editor = {IEEE},
  month = may,
  pages = {382-387 vol.1},
  url = {https://doi.org/10.1109/ROBOT.1993.292011},
  urldate = {2025-05-07}
}

@book{hamannSwarmRoboticsFormal2018,
  title = {Swarm {{Robotics}}: {{A Formal Approach}}},
  shorttitle = {Swarm {{Robotics}}},
  author = {Hamann, Heiko},
  year = {2018},
  publisher = {Springer International Publishing},
  address = {Cham},
  url = {https://doi.org/10.1007/978-3-319-74528-2},
  urldate = {2025-07-01},
  copyright = {http://www.springer.com/tdm},
  isbn = {978-3-319-74526-8 978-3-319-74528-2},
  langid = {english}
}

@article{capPrioritizedPlanningAlgorithms2015a,
  title = {Prioritized {{Planning Algorithms}} for {{Trajectory Coordination}} of {{Multiple Mobile Robots}}},
  author = {{\v C}{\'a}p, Michal and Nov{\'a}k, Peter and Kleiner, Alexander and Seleck{\'y}, Martin},
  year = {2015},
  month = jul,
  journal = {IEEE Transactions on Automation Science and Engineering},
  volume = {12},
  number = {3},
  pages = {835--849},
  issn = {1558-3783},
  url = {https://doi.org/10.1109/TASE.2015.2445780},
  urldate = {2025-07-03}
}

@article{sternMultiAgentPathfindingDefinitions2019,
  title = {Multi-{{Agent Pathfinding}}: {{Definitions}}, {{Variants}}, and {{Benchmarks}}},
  shorttitle = {Multi-{{Agent Pathfinding}}},
  author = {Stern, Roni and Sturtevant, Nathan and Felner, Ariel and Koenig, Sven and Ma, Hang and Walker, Thayne and Li, Jiaoyang and Atzmon, Dor and Cohen, Liron and Kumar, T. K. and Bart{\'a}k, Roman and Boyarski, Eli},
  year = {2019},
  journal = {Proceedings of the International Symposium on Combinatorial Search},
  volume = {10},
  number = {1},
  pages = {151--158},
  issn = {2832-9163},
  url = {https://doi.org/10.1609/socs.v10i1.18510},
  urldate = {2025-07-03},
  copyright = {Copyright (c) 2021 Proceedings of the International Symposium on Combinatorial Search},
  langid = {english}
}

@book{engelbrechtFundamentalsComputationalSwarm2006,
  title = {Fundamentals of {{Computational Swarm Intelligence}}},
  author = {Engelbrecht, Andries P.},
  year = {2006},
  publisher = {John Wiley \& Sons, Inc.},
  address = {Hoboken, NJ, USA},
  isbn = {978-0-470-09191-3}
}

@incollection{frayContextSteeringBehaviorDriven2019,
  title = {Context {{Steering}}: {{Behavior-Driven Steering}} at the {{Macro Scale}}},
  shorttitle = {Context {{Steering}}},
  booktitle = {Game {{AI Pro}} 360: {{Guide}} to {{Movement}} and {{Pathfinding}}},
  author = {Fray, Andrew},
  year = {2019},
  editor = {Game AI Pro 360},
  publisher = {CRC Press},
  isbn = {978-0-429-05509-6},
  url = {https://www.taylorfrancis.com/chapters/edit/10.1201/9780429055096-14/context-steering-andrew-fray}
}

@inproceedings{dockhornMultiObjectiveOptimizationDecisionMaking2021,
  title = {Multi-{{Objective Optimization}} and {{Decision-Making}} in {{Context Steering}}},
  booktitle = {2021 {{IEEE Conference}} on {{Games}} ({{CoG}})},
  author = {Dockhorn, Alexander and Mostaghim, Sanaz and Kirst, Martin and Zettwitz, Martin},
  year = {2021},
  editor = {IEEE},
  month = aug,
  pages = {1--8},
  issn = {2325-4289},
  url = {https://doi.org/10.1109/CoG52621.2021.9619155},
  urldate = {2025-09-02}
}

@mastersthesis{thoms_anwendung_2021,
	title = {Anwendung von dreidimensionalem {Context}-{Steering} auf {Quadcopter}-{Schwärme}},
	school = {OVGU Magdeburg},
	author = {Thoms, Philipp},
	month = sep,
	year = {2021},
    url = {https://www.ci.ovgu.de/is_media/Master+und+Bachelor_Arbeiten+/Master_PhilippThoms_pdf-p-7562.pdf},
}

@mastersthesis{stein_multikriteriell_2018,
	title = {Multikriteriell optimiertes {Context} {Steering} für autonome {Bewegung} im {Gebiet} der {Schwarmrobotik}},
	language = {de},
	school = {OVGU Magdeburg},
	author = {Stein, Andrei},
	month = jan,
	year = {2018},
    url = {https://ci.ovgu.de/is_media/Master+und+Bachelor_Arbeiten+/MAStein.pdf},
}

@article{thrunStanleyRobotThat2006,
  title = {Stanley: {{The}} Robot That Won the {{DARPA Grand Challenge}}},
  shorttitle = {Stanley},
  author = {Thrun, Sebastian and Montemerlo, Mike and Dahlkamp, Hendrik and Stavens, David and Aron, Andrei and Diebel, James and Fong, Philip and Gale, John and Halpenny, Morgan and Hoffmann, Gabriel and Lau, Kenny and Oakley, Celia and Palatucci, Mark and Pratt, Vaughan and Stang, Pascal and Strohband, Sven and Dupont, Cedric and Jendrossek, Lars-Erik and Koelen, Christian and Markey, Charles and Rummel, Carlo and {van Niekerk}, Joe and Jensen, Eric and Alessandrini, Philippe and Bradski, Gary and Davies, Bob and Ettinger, Scott and Kaehler, Adrian and Nefian, Ara and Mahoney, Pamela},
  year = {2006},
  journal = {Journal of Field Robotics},
  volume = {23},
  number = {9},
  pages = {661--692},
  issn = {1556-4967},
  url = {https://doi.org/10.1002/rob.20147},
  urldate = {2025-09-03},
  langid = {english}
}

@article{sniderAutomaticSteeringMethods2009,
  title = {Automatic Steering Methods for Autonomous Automobile Path Tracking},
  author = {Snider, Jarrod M.},
  year = {2009},
  journal = {Robotics Institute, Pittsburgh, PA, Tech. Rep. CMU-RITR-09-08},
  urldate = {2025-09-03},
  url = {https://publications.ri.cmu.edu/storage/publications/pub_files/2009/2/Automatic_Steering_Methods_for_Autonomous_Automobile_Path_Tracking.pdf}
}

@inproceedings{schonnagelAVOIDJACKAvoidanceJackknifing2025,
  author={Schönnagel, Adrian and Dubé, Michael and Steup, Christoph and Keppler, Felix and Mostaghim, Sanaz},
  booktitle={2025 IEEE International Symposium on Multi-Robot and Multi-Agent Systems (MRS)}, 
  title={AVOID-JACK: Avoidance of Jackknifing for Swarms of Long Heavy Articulated Vehicles}, 
  year={2025},
  editor={IEEE},
  volume={},
  number={},
  pages={1-7},
  url={https://doi.org/10.1109/MRS66243.2025.11357246}
}

@article{dmpc_truck,
    author = {Dai, Li and Hao, Yanye and Xie, Huahui and Sun, Zhongqi and Xia, Yuanqing},
    title = {Distributed robust {MPC} for nonholonomic robots with obstacle and collision avoidance},
    journal = {Control Theory and Technology},
    volume = 20,
    pages = {32--45},
    year = 2022,
    publisher = {Springer Nature},
    url = {https://doi.org/10.1007/s11768-022-00079-x}
}

@article{dmpc_truck_trailer_platooning,
    title = {Enhancing truck platooning efficiency and safety—{A} distributed {Model Predictive Control} approach for lane-changing manoeuvres},
    journal = {Control Engineering Practice},
    volume = {154},
    pages = {106153},
    year = {2025},
    url = {https://doi.org/10.1016/j.conengprac.2024.106153},
    author = {Beatriz Lourenço and Daniel Silvestre},
    publisher = {Elsevier}
}

@ARTICLE{mrs_distributed_control,
  author={Jin, Zengke and Wang, Chaoli and Liang, Dong and Wang, Shanshan and Ding, Zhengtao},
  journal={IEEE Transactions on Intelligent Vehicles}, 
  title={Fixed-Time Consensus for Multiple Tractor-Trailer Vehicles With Dynamics Control: A Distributed Internal Model Approach}, 
  year={2024},
  volume={9},
  number={1},
  pages={656-669},
  url = {https://doi.org/10.1109/TIV.2023.3338238},
}

@article{parzenKDE1962,
author = {Emanuel Parzen},
title = {{On Estimation of a Probability Density Function and Mode}},
volume = {33},
journal = {The Annals of Mathematical Statistics},
number = {3},
publisher = {Institute of Mathematical Statistics},
pages = {1065 -- 1076},
year = {1962},
url = {https://doi.org/10.1214/aoms/1177704472}
}

@article{rosenblattKDE1956,
author = {Murray Rosenblatt},
title = {{Remarks on Some Nonparametric Estimates of a Density Function}},
volume = {27},
journal = {The Annals of Mathematical Statistics},
number = {3},
publisher = {Institute of Mathematical Statistics},
pages = {832 -- 837},
year = {1956},
url = {https://doi.org/10.1214/aoms/1177728190}
}

@misc{zenodoANTS2025,
  author       = {{Schönnagel, Adrian and
                  Steup, Christoph and
                  Dube, Michael and
                  Keppler, Felix and
                  Mostaghim, Sanaz}},
  title        = {{Supplementary material for "PREVENT-JACK: Context
                   Steering for Swarms of Long Heavy Articulated
                   Vehicles"}},
  month        = nov,
  year         = 2025,
  publisher    = {Zenodo},
  doi          = {10.5281/zenodo.17650869},
  url          = {https://doi.org/10.5281/zenodo.17650869}
}

@INPROCEEDINGS{kottingerKCBS,
  author={Kottinger, Justin and Almagor, Shaull and Lahijanian, Morteza},
  booktitle={2022 IEEE/RSJ International Conference on Intelligent Robots and Systems (IROS)}, 
  title={Conflict-Based Search for Multi-Robot Motion Planning with Kinodynamic Constraints}, 
  year={2022},
  editor = {IEEE},
  volume={},
  number={},
  pages={13494-13499},
  url={https://doi.org/10.1109/IROS47612.2022.9982018}
}

@INPROCEEDINGS{birdDvmSlam25,
  author={Bird, Joshua and Blumenkamp, Jan and Prorok, Amanda},
  booktitle={2025 IEEE International Conference on Robotics and Automation (ICRA)}, 
  title={DVM-SLAM: Decentralized Visual Monocular Simultaneous Localization and Mapping for Multi-Agent Systems}, 
  year={2025},
  editor = {IEEE},
  volume={},
  number={},
  pages={1-7},
  url = {https://doi.org/10.1109/ICRA55743.2025.11127510}}

@ARTICLE{lajoieSwarmSlam24,
  author={Lajoie, Pierre-Yves and Beltrame, Giovanni},
  journal={IEEE Robotics and Automation Letters}, 
  title={Swarm-SLAM: Sparse Decentralized Collaborative Simultaneous Localization and Mapping Framework for Multi-Robot Systems}, 
  year={2024},
  volume={9},
  number={1},
  pages={475-482},
  url = {https://doi.org/10.1109/LRA.2023.3333742}}

@INPROCEEDINGS{hornyaUvdar22,
  author={Horyna, Jiri and Walter, Viktor and Saska, Martin},
  booktitle={2022 International Conference on Unmanned Aircraft Systems (ICUAS)}, 
  title={UVDAR-COM: UV-Based Relative Localization of UAVs with Integrated Optical Communication}, 
  year={2022},
  editor = {ICUAS},
  volume={},
  number={},
  pages={1302-1308},
  url = {https://doi.org/10.1109/ICUAS54217.2022.9836151}}

@ARTICLE{walterUvdar19,
  author={Walter, Viktor and Staub, Nicolas and Franchi, Antonio and Saska, Martin},
  journal={IEEE Robotics and Automation Letters}, 
  title={UVDAR System for Visual Relative Localization With Application to Leader–Follower Formations of Multirotor UAVs}, 
  year={2019},
  volume={4},
  number={3},
  pages={2637-2644},
  url = {https://doi.org/10.1109/LRA.2019.2901683}}

@INPROCEEDINGS{vandenBergRVO08,
  author={van den Berg, Jur and Ming Lin and Manocha, Dinesh},
  booktitle={2008 IEEE International Conference on Robotics and Automation}, 
  title={Reciprocal Velocity Obstacles for real-time multi-agent navigation}, 
  year={2008},
  editor = {IEEE},
  volume={},
  number={},
  pages={1928-1935},
  url = {https://doi.org/10.1109/ROBOT.2008.4543489}}

@InProceedings{vandenBergORCA11,
    author="van den Berg, Jur
    and Guy, Stephen J.
    and Lin, Ming
    and Manocha, Dinesh",
    editor="Pradalier, C{\'e}dric
    and Siegwart, Roland
    and Hirzinger, Gerhard",
    title="Reciprocal n-Body Collision Avoidance",
    booktitle="Robotics Research",
    year="2011",
    publisher="Springer Berlin Heidelberg",
    address="Berlin, Heidelberg",
    pages="3--19",
    isbn="978-3-642-19457-3",
    url = {https://link.springer.com/chapter/10.1007/978-3-642-19457-3_1}
}

@incollection{alonso-moraOptimalReciprocalCollision2013,
  title = {Optimal {{Reciprocal Collision Avoidance}} for {{Multiple Non-Holonomic Robots}}},
  booktitle = {Distributed {{Autonomous Robotic Systems}}: {{The}} 10th {{International Symposium}}},
  author = {{Alonso-Mora}, Javier and Breitenmoser, Andreas and Rufli, Martin and Beardsley, Paul and Siegwart, Roland},
  editor = {Martinoli, Alcherio and Mondada, Francesco and Correll, Nikolaus and Mermoud, Gr{\'e}gory and Egerstedt, Magnus and Hsieh, M. Ani and Parker, Lynne E. and St{\o}y, Kasper},
  year = 2013,
  pages = {203--216},
  publisher = {Springer},
  address = {Berlin, Heidelberg},
  url = {https://doi.org/10.1007/978-3-642-32723-0_15},
  urldate = {2026-03-30},
  isbn = {978-3-642-32723-0},
  langid = {english}
}

@article{bareissGeneralizedReciprocalCollision2015,
  title = {Generalized Reciprocal Collision Avoidance},
  author = {Bareiss, Daman and {van den Berg}, Jur},
  year = 2015,
  month = oct,
  journal = {The International Journal of Robotics Research},
  volume = {34},
  number = {12},
  pages = {1501--1514},
  publisher = {SAGE Publications Ltd STM},
  issn = {0278-3649},
  url = {https://doi.org/10.1177/0278364915576234},
  urldate = {2026-03-30},
  langid = {english}
}

@article{rufliReciprocalCollisionAvoidance2013,
  title = {Reciprocal {{Collision Avoidance With Motion Continuity Constraints}}},
  author = {Rufli, Martin and {Alonso-Mora}, Javier and Siegwart, Roland},
  year = 2013,
  month = aug,
  journal = {IEEE Transactions on Robotics},
  volume = {29},
  number = {4},
  pages = {899--912},
  issn = {1941-0468},
  url = {https://doi.org/10.1109/TRO.2013.2258733},
  urldate = {2026-03-30}
}

@article{maoOrcaMPC20,
author = {Mao, Run and Gao, Hongli and Guo, Liang},
title = {A Novel Collision-Free Navigation Approach for Multiple Nonholonomic Robots Based on ORCA and Linear MPC},
journal = {Mathematical Problems in Engineering},
volume = {2020},
number = {1},
pages = {4183427},
url = {https://onlinelibrary.wiley.com/doi/abs/10.1155/2020/4183427},
year = {2020}
}

@INPROCEEDINGS{bone23MonteCarlo,
  author={Bone, Sean and Bartolomei, Luca and Kennel-Maushart, Florian and Chli, Margarita},
  booktitle={2023 IEEE/RSJ International Conference on Intelligent Robots and Systems (IROS)}, 
  title={Decentralised Multi-Robot Exploration Using Monte Carlo Tree Search}, 
  year={2023},
  volume={},
  number={},
  publisher={IEEE},
  editor={IEEE},
  pages={7354-7361},
  url = {https://doi.org/10.1109/IROS55552.2023.10341485}
}

@Article{Salamah23TruckReversing,
AUTHOR = {Bin Salamah, Yasser},
TITLE = {Sliding Mode Controller for Autonomous Tractor-Trailer Vehicle Reverse Path Tracking},
JOURNAL = {Applied Sciences},
VOLUME = {13},
YEAR = {2023},
NUMBER = {21},
ARTICLE-NUMBER = {11998},
ISSN = {2076-3417},
url = {https://doi.org/10.3390/app132111998}
}

@inproceedings{perez13RHEA,
author = {Perez, Diego and Samothrakis, Spyridon and Lucas, Simon and Rohlfshagen, Philipp},
title = {Rolling horizon evolution versus tree search for navigation in single-player real-time games},
year = {2013},
isbn = {9781450319638},
publisher = {Association for Computing Machinery},
editor = {ACM},
address = {New York, NY, USA},
url = {https://doi.org/10.1145/2463372.2463413},
booktitle = {Proceedings of the 15th Annual Conference on Genetic and Evolutionary Computation},
pages = {351–358},
numpages = {8},
location = {Amsterdam, The Netherlands},
series = {GECCO '13}
}

@ARTICLE{gaina22RHEA,
  author={Gaina, Raluca D. and Devlin, Sam and Lucas, Simon M. and Perez-Liebana, Diego},
  journal={IEEE Transactions on Games}, 
  title={Rolling Horizon Evolutionary Algorithms for General Video Game Playing}, 
  year={2022},
  volume={14},
  number={2},
  pages={232-242},
  url = {https://doi.org/10.1109/TG.2021.3060282}}

\end{document}